\documentclass[iicol,sn-basic]{sn-jnl}

\newtheorem{Theorem}{Theorem}
\usepackage{graphicx}%
\usepackage{multirow}%
\usepackage{amsmath,amssymb,amsfonts}%
\usepackage{amsthm}%
\usepackage{mathrsfs}%
\usepackage[title]{appendix}%
\usepackage{xcolor}%
\usepackage{textcomp}%
\usepackage{manyfoot}%
\usepackage{booktabs}%
\usepackage{algorithm}%
\usepackage{algorithmicx}%
\usepackage{algpseudocode}%
\usepackage{listings}%
\definecolor{darkgreen}{rgb}{0.0, 0.5, 0.0}
\usepackage{colortbl}
\usepackage{hhline}
\usepackage{multirow}
\usepackage[table,xcdraw]{xcolor}
\usepackage{graphicx} 
\usepackage{adjustbox} 
\usepackage{wrapfig} 
\usepackage[normalem]{ulem}
\usepackage{arydshln}
\usepackage{amsmath}  
\usepackage{amsthm}
\usepackage{graphicx}
\usepackage{tikz}
\usepackage{comment}
\usepackage{amsmath,amssymb} 
\usepackage{multirow}
\usepackage{multirow}
\usepackage{algorithm}  
\usepackage{algorithmicx}  
\usepackage{algpseudocode}  
\usepackage{array}  
 
\usepackage[utf8]{inputenc} 
\usepackage[T1]{fontenc}    
\usepackage{hyperref}       
\usepackage{url}            
\usepackage{booktabs}       
\usepackage{amsfonts}       
\usepackage{nicefrac}       
\usepackage{microtype}      
\usepackage{algorithm}  
\usepackage{algorithmicx}  
\usepackage{algpseudocode}  
\usepackage{wrapfig} 
\usepackage{lipsum}  
\usepackage{amsmath}
\usepackage{multirow}       
\usepackage{amsthm}
\usepackage{graphicx}

\usepackage{footmisc}
\usepackage{amsthm}

\theoremstyle{thmstyleone}%
\newtheorem{corollary}{Corollary}

\theoremstyle{thmstyletwo}%

\theoremstyle{thmstylethree}%

\usepackage{graphicx}
\begin{document}

\title[Article Title]{Knowledge-Guided Adversarial Training for Infrared Object Detection via Thermal Radiation Modeling}


\author[1]{Shiji Zhao}\email{zhaoshiji123@buaa.edu.cn}
\equalcont{These authors contributed equally to this work.}

\author[1]{Shukun Xiong}\email{shukunxiong@buaa.edu.cn}
\equalcont{These authors contributed equally to this work.}

\author[1]{Maoxun Yuan}\email{yuanmaoxun@buaa.edu.cn}

\author[1]{Yao Huang}\email{y\_huang@buaa.edu.cn}

\author[2]{Ranjie Duan}\email{ranjieduan@gmail.com}

\author[3]{Qing Guo}\email{tsingqguo@ieee.org}

\author[4]{Jiansheng Chen}\email{schen@ustb.edu.cn}

\author[5]{Haibin Duan}\email{hbduan@buaa.edu.cn}

\author*[1]{Xingxing Wei}\email{xxwei@buaa.edu.cn}


\affil[1]{Institute of Artificial Intelligence, Beihang University, No.37 Xueyuan Road, Haidian District, Beijing 100191, P.R. China}
\affil[2]{Security Department, Alibaba Group, Hangzhou 310056, China}
\affil[3]{School of Computer Science, Nankai University, China}
\affil[4]{School of Computer and Communication Engineering, University of Science and Technology Beijing, Beijing 100083, China}
\affil[5]{School of Automation Science and Electrical Engineering, Beihang University, Beijing 100083, China}


\abstract{In complex environments, infrared object detection exhibits broad applicability and stability across diverse scenarios.  However, infrared object detection is vulnerable to both common corruptions and adversarial examples, leading to potential security risks. To improve the robustness of infrared object detection, current methods mostly adopt a data-driven ideology, which only superficially drives the network to fit the training data without specifically considering the unique characteristics of infrared images, resulting in limited robustness. In this paper, we revisit infrared physical knowledge and find that relative thermal radiation relations between different classes can be regarded as a reliable knowledge source under the complex scenarios of adversarial examples and common corruptions. Thus, we theoretically model thermal radiation relations based on the rank order of gray values for different classes, and further quantify the stability of various inter-class thermal radiation relations. Based on the above theoretical framework, we propose Knowledge-Guided Adversarial Training (KGAT) for infrared object detection, in which infrared physical knowledge is embedded into the adversarial training process, and the predicted results are optimized to be consistent with the actual physical laws. \textcolor{black}{Extensive experiments on three infrared datasets and six mainstream infrared object detection models demonstrate that KGAT effectively enhances both clean accuracy and robustness against adversarial attacks and common corruptions.}}




\maketitle

\section{Introduction}\label{sec1}

Compared with RGB images, infrared images play an irreplaceable role in security surveillance \citep{suard2006pedestrian}, autonomous driving, and remote sensing detection \citep{weng2009thermal}, as well as other fields, owing to their excellent environmental adaptability and robust anti-interference performance. These advantages enable infrared imaging to maintain high performance in adverse conditions, such as low-light, smoky, and other visually obscured scenarios. With the rapid development of computer vision, infrared images are widely applied in different tasks, e.g., object detection \citep{lin2023learning,liu2023infrared} and semantic segmentation \citep{li2023near,tian2023vu}.

However, due to its wide application in complex scenarios, infrared object detection is exposed to considerable security risks under adversarial attacks \citep{madry2017towards,zhu2022infrared, zhu2021fooling, wei2023physically,carlini2017cw,madry2018pgd,rakin2019mtd} and common corruptions (e.g., salt noise, Gaussian blur). Current robustness enhancement techniques focus more on visible light scenes \citep{rakin2019mtd,Chen2024Accurate,Dong2022AdversariallyAware,zhao2024mitigating,wei2024revisiting}. Although these methods can  be adapted to infrared modalities with infrared data, they often yield limited robustness gains while compromising clean accuracy, motivating further investigation.. 

Here, we conduct an in-depth analysis of the causes of insufficient robustness in infrared object detection. We find that most existing methods for enhancing the robustness of infrared object detection adopt a data-driven approach: the trained model only superficially fits the infrared data yet ignores the intrinsic exploration towards unique infrared physical knowledge, resulting in poor generalization on out-of-distribution data  and vulnerability to adversarial attacks and common corruptions. Meanwhile, current knowledge-guided visual recognition methods \citep{takikawa2019gated, li2023recognizing, sitawarin2022part,gurel2021knowledge, li2023recognizing, zhang2023care} also lack the corresponding research specifically for infrared scenes.
Thus, it is imperative to explicitly model the infrared physical knowledge and further enhance the robustness against adversarial attacks and common corruptions.


\begin{figure}
    \centering
    \includegraphics[width=0.98\linewidth]{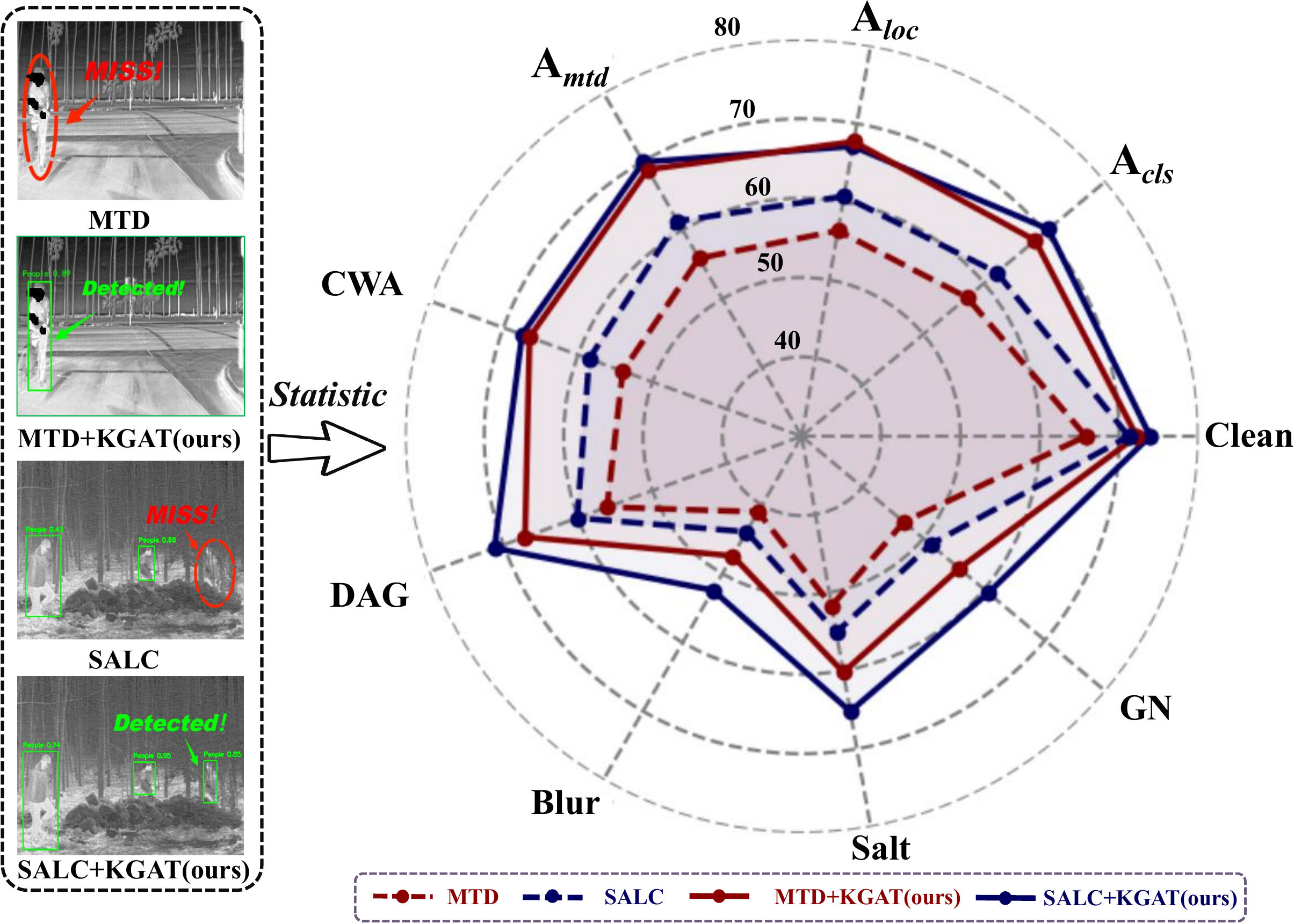}
    \vspace{-3pt}
    \caption{\textbf{Performance of the baseline and our knowledge-guided adversarial training (KGAT) on M$^{3}$FD dataset, using YOLO-v8 as the detector.} Compared with Data-driven method,  KGAT can enhance the robustness of infrared object detection towards more diverse perturbations in complex infrared environments.}
    \label{fig:enter-label}
\end{figure}

To enhance the robustness of infrared object detectors, we should analyze what infrared physical knowledge is available and robust. We initially consider the thermal radiation characteristic: in contrast to RGB images, the most prominent feature of infrared images is that their gray values are determined by the thermal radiation characteristics of object categories, e.g., material properties \citep{grossmann2022improving}.  Based on the blackbody radiation rule, every object can be theoretically modeled by its unique thermal radiation characteristic at a known temperature. However, the direct application of the blackbody radiation law poses numerous challenges in practical scenarios. In real-world scenarios, environmental factors, e.g., illumination, temperature, and humidity, complicate the gray values of classes in images and their thermophysical properties. Additionally, under adversarial attacks \citep{wei2023physically,rakin2019mtd,madry2018pgd} and common corruptions like salt noise , Gaussian noise, and Gaussian  blur, distinct correlation patterns may emerge, making it difficult to directly establish a general mathematical framework for the infrared physical knowledge.

\begin{figure}[t]
    \centering
    \includegraphics[trim=0cm 0cm 0cm 0cm, clip,width=\linewidth]{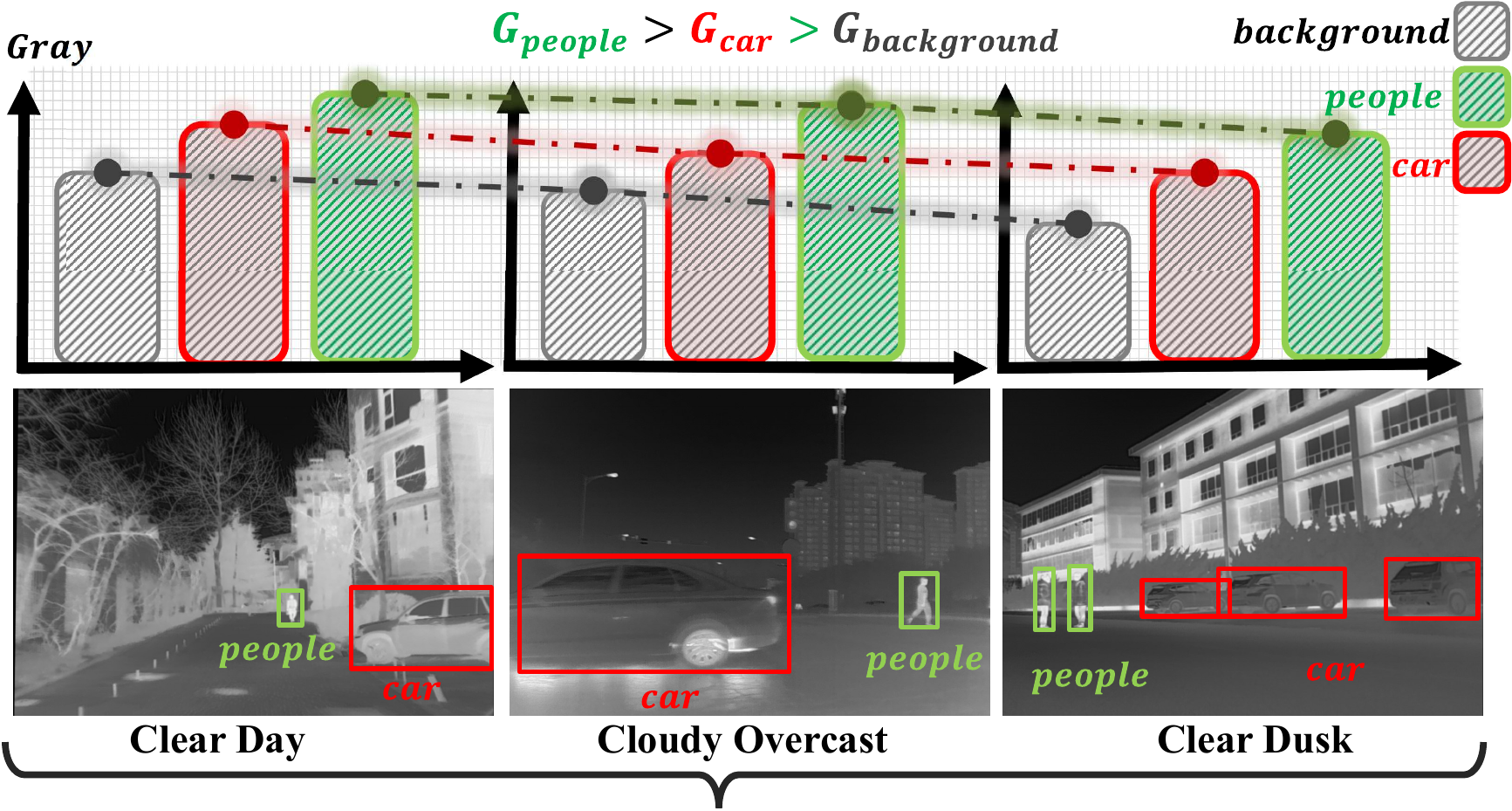}
    \vspace{-20pt}
    \caption{\textbf{The advantage of relative thermal radiation relations.} In the M$^3$FD dataset, under environmental conditions such as Clear Day, Cloudy Overcast, and Clear Dusk, although the absolute gray values of the ``car'' and the ``people'' are different, most of the relative thermal radiation relations between ``car'' and the ``people'' keep stable.}
    \label{fig:infrared image}
\end{figure}

Fortunately, as shown in Figure \ref{fig:infrared image}, under different environmental conditions, although the absolute gray values of various classes may vary, the thermal radiation relations between certain classes remain stable in most images. This principle  also holds when the classes are subjected to perturbations within a certain range. Therefore, these relations can be regarded as relatively reliable sources of infrared physical knowledge and have the potential to enhance the robustness for infrared object detection. To apply thermal radiation relations in object detection, two obvious challenges need to be solved: \textbf{First, it is challenging to  directly model the thermal radiation relations in the optimization of infrared object detection.} The relative gray value relation is an intuitive  concept, but how to mathematically model the relative gray value relations among all classes in an image remains an open question. At the same time, how to transform the concept of relative gray value relation into an actual optimization goal that can guide the model training process also needs to be solved. \textbf{Second, not all thermal radiation relations are stable.} Since the thermal radiation relations are affected by external factors (such as temperature fluctuations and weather conditions),  some thermal radiation relations are unstable and variable in different scenes. Hence, a rigorous metric is required to quantify the stability of  thermal radiation relations. Relations proven to be relatively reliable merit greater attention, whereas those that waver should be approached with more caution.

To address these two challenges, we propose a novel theoretical framework to model the thermal radiation relation and further quantify their stability. Specifically, \textbf{to theoretically model thermal radiation relation}, we utilize the rank order of the gray value relation between different classes to reflect the relative thermal radiation relation in a single image. We also employ the Spearman rank correlation coefficient to measure the discrepancy between predicted results and the actual thermal radiation relation, which can transform the concept of the relation into concrete mathematical variables and can be further regarded as the optimization goal for the infrared object detection in the training process. \textbf{To quantify the stability of the thermal radiation relation}, we count all the variations of the relative thermal radiation relation between different classes in the entire scenes. 
Consequently, a smaller variation indicates a more stable intrinsic physical constraint, which should increase the corresponding optimization strength for this type of infrared physical knowledge.

To utilize the modeled infrared physical knowledge, we design a Knowledge-Guided Adversarial Training  \textbf{KGAT} method for infrared object detection based on the relative thermal radiation characteristic. Specifically, KGAT initially extracts the Spearman rank correlation coefficient of the thermal radiation relation between the predicted result and ground truth. Since the correlation coefficient for infrared knowledge is difficult to derive in the actual optimization process, KGAT transforms direct constraints for the prediction results of thermal radiation relation into indirectly adjusting the detection loss weight, which is theoretically proven to be effective to achieve the final optimization goal. In addition, KGAT extracts the variation of the thermal radiation relation in the entire scenes and applies it to adjust the optimization strength for the infrared physical knowledge, so that the model will focus more on the stable thermal radiation relation in the training process. A series of experiments show that our KGAT can not only effectively improve the clean accuracy, but can also enhance the robustness against adversarial attacks (e.g., PGD and adversarial patch), and common corruptions (e.g., salt  noise and blurring). Our code is available at \url{https://github.com/shukunxiong/KGAT}.

Our contribution can be summarized as follows:

\begin{itemize}
\item We introduce a novel paradigm that integrates knowledge-guided principles with original data-driven techniques for infrared object detection. To fully utilize the infrared physical knowledge, we develop a theoretical framework to model the thermal radiation relation and further measure their stability.
\item We design a Knowledge-Guided Adversarial Training method (KGAT) for infrared object detection, which utilizes the thermal radiation relation as an  indirect constraint to guide the model prediction results to match the real infrared physical knowledge.
\item A series of experiments show that KGAT can enhance the accuracy and robustness against different types of common corruptions and adversarial attacks in three infrared datasets and six object detections. In particular, KGAT achieves an average robustness improvement of 9.0\% over the second-best method on DINO of FLIR-ADAS.
\end{itemize}

The rest of the paper is organized as follows: Related work is given in Section \ref{sec:related work}. Section \ref{sec:motivation} introduces the theoretical framework for the thermal radiation relation. Section \ref{sec:method} introduces the details of our knowledge-guided adversarial training method. The experiments are conducted in Section \ref{sec:experiment}, the limitations are listed in Section \ref{sec:limitation}, and the conclusion is given in Section \ref{sec:Conclusion}.

\section{Related Work}
\label{sec:related work}
\subsection{Infrared Object Detection}

\textcolor{black}{
Since visible modality suffers from limited informative content under low-illumination conditions, infrared modality has been utilized as an alternative to achieve a full-day object detection task. Some methods~\citep{chen2018infrared, li2021yolo, dai2021tirnet, li2022dense, zhang2022isnet} have been proposed to improve the performance of infrared object detection.
\cite{chen2018infrared} introduce the top and bottom-hat transformation to enhance object outlines and then extracts Haar-like features for detection. YOLO-FIR~\citep{li2021yolo} is proposed as a region-free detector for weak objects in infrared images. Moreover, TIRNet~\citep{dai2021tirnet} achieves reliable and efficient infrared detection for autonomous driving. For infrared small object detection, ACM embeds low-level details into high-level features, DNANet~\citep{li2022dense} leverages contextual information via feature fusion, and ISNet~\citep{zhang2022isnet} incorporates deformable convolutions to enhance performance.}

\textcolor{black}{
Recent advances further develop specialized infrared architectures to address low contrast, thermal noise, and domain-specific cues, especially for infrared small target detection. For example, RKformer~\citep{zhang2022rkformer} strengthens long-range dependency modeling while preserving fine details, Dim2Clear~\citep{zhang2023dim2clear} enhances dim and tiny targets under cluttered thermal backgrounds, IRPruneDet~\citep{zhang2024irprunedet} improves efficiency via structured regularization and channel pruning, and IRSAM~\citep{zhang2024irsam} adapts SAM-style representations to capture subtle boundaries in infrared imagery. Beyond detection, IRGPT~\citep{cao2025irgpt} explores large-scale infrared vision-language understanding, diffusion-based visible-infrared fusion such as Dif-CDFusion~\citep{liu2025difcdfusion} exploits complementary spectral cues, and real-world label noise has been explicitly studied in visible-infrared re-identification~\citep{zhang2025vireidnoise}, highlighting the practical challenge of imperfect supervision.
}

\textcolor{black}{
Unlike the aforementioned methods that mainly redesign task-specific architectures or focus on texture or shape cues in infrared images, our knowledge-guided training method constrains the model predictions to be consistent with the inherent infrared thermal radiation relations, thereby improving adversarial robustness and remaining complementary to various infrared detectors.
}

\subsection{Adversarial Attacks for Infrared Object Detection}

Although infrared object detection has demonstrated promising performance under low-illumination conditions, several studies have found that these models exhibit limited robustness to various corruptions, particularly adversarial attacks. Specifically, \cite{zhu2021fooling} first demonstrate the vulnerability of infrared detectors by designing heat-emitting adversarial bulbs and further design code-patterned thermal-masking clothing \citep{zhu2022infrared}, proving that infrared detectors lack robustness against some specific patterns that deliberately manipulate thermal radiation. Furthermore, Wei et al. highlight these robustness concerns by demonstrating that simple infrared adversarial patches \citep{wei2023physically, wei2023unified} can successfully attack both pedestrian and vehicle detectors, emphasizing the necessity of enhancing the robustness of current infrared detectors. 

\subsection{Adversarial Robustness in Infrared Object Detection}

For the adversarial training for object detection, current methods focus more on enhancing the adversarial robustness for RGB images \citep{zhang2019mtd, Chen2024Accurate,Dong2022AdversariallyAware,xu2022robust, chen2021class}. \cite{zhang2019mtd} analyze the loss functions of object detection methods from a multi-task perspective and leverage  multiple sources of attacks to enhance the robustness of detection models. \cite{Dong2022AdversariallyAware} propose a Robust Detector (RobustDet) based on adversarially aware convolution to disentangle gradients for model learning on both clean and adversarial images. \cite{chen2021class} apply class-weighted loss to generate corresponding adversarial examples and further enhance the adversarial robustness for object detection. \cite{Chen2024Accurate} apply the adversarial examples of model self-evolution to dynamically enhance the detector performance. \cite{xu2022robust} propose unified decoupled feature alignment to utilize self-distillation to enhance the adversarial robustness of object detection models. However, these methods do not account for the infrared modality, leading to insufficient adversarial robustness. Several studies try to enhance the robustness of the infrared detectors. For example, Patch-based Occlusion-Aware Detection (POD) \citep{strack2024defending}  augments training with random patches and the diffusion-based approach DIFFender \citep{wei2024real} localizes and restores adversarial regions in infrared images.

Unlike the aforementioned methods, our knowledge-guided adversarial training guides the network to learn the intrinsic thermal radiation principles, which can effectively enhance adversarial robustness in the training process rather than treating it as a separate defensive module, thus avoiding the inherent limitations of previous methods.

\subsection{Knowledge-Guided Visual Recognition}

Different from data-driven methods, knowledge-guided methods are often  in line with human visual cognition of the physical world and can enhance the model's interpretability and robustness. Previous knowledge-guided work on improving visual recognition attempts to use shape knowledge \citep{takikawa2019gated, li2023recognizing, sitawarin2022part}, logical knowledge \citep{gurel2021knowledge, li2023recognizing, zhang2023care}, and causal knowledge \citep{zhang2021causaladv, kim2023demystifying, zhang2020causal}. As for the shape knowledge, it has been widely applied to various aspects of visual recognition tasks.  \cite{takikawa2019gated} apply gated convolution to extract shape and contour features and combine it with traditional image gradient information (the image gradient extracted by the Sobel operator) to obtain contour information to improve segmentation results and further enhance object detection performance. \cite{sitawarin2022part} combines a part segmentation model with a lightweight classifier and trains it end-to-end to simultaneously segment objects into parts and then classify the segmented objects to enhance the adversarial robustness. 
For the logic knowledge, \cite{gurel2021knowledge} first integrate domain knowledge into a probabilistic graphical model via first-order logic rules in recognizing traffic signs, and \cite{zhang2023care} extend it into more general visual recognition tasks. \cite{li2023recognizing} utilize both the shape and logical knowledge, and applies different part-based segmentation models to predict the part and apply human prior logical knowledge towards different shapes to judge the final predicted classes. As for causal knowledge, \cite{zhang2020causal} analyze the robustness of DNNs against input manipulations via a causal perspective. \cite{zhang2021causaladv} further describe causal relation in adversarial situations and applies them to enhance the robustness of adversarial training. \cite{kim2023demystifying} utilize instrumental variable regression to generate robust features that are free of confounding effects.
 
However, the above methods are all knowledge-guided methods for visible light scenes and remain under-explored in infrared object detection. In this work, we fully exploit the relative thermal radiation relation as a form of infrared physical knowledge in infrared images and use it to enhance the robustness of infrared object detection.

\section{Thermal Radiation Relation}
\label{sec:motivation}

\subsection{Modeling of Thermal Radiation Relation}

\textcolor{black}{Based on the principles of infrared thermography, we first connect thermal radiance to class-wise mean gray values, and then model inter-class thermal radiation relations via rank order. For an infrared image $x$ representing a scene with $K$ distinct classes, we establish Corollary \ref{corollary 1}  based on Planck’s law of blackbody spectral radiance \citep{Modest2013RadiativeHeatTransfer,Planck1901NormalSpectrum} to describe how intrinsic radiation is captured by the imaging pipeline as gray values:}
\textcolor{black}{
\begin{corollary}
\label{corollary 1}
Consider the annotated region $\Omega_k^{(i)}$ of the $k$-th class in the $i$-th infrared image.
Let $\bar L_k(u)$ denote the effective radiance at pixel $u$ contributed by the object of the $k$-th class, $\delta(u)$ denotes the class-agnostic gray value estimation error at location $u$, which arises from environment noise or adversarial perturbation,
and $\mathbb{E}_{u\sim\Omega_k^{(i)}}[\cdot]$ denote the uniform spatial average over pixel locations $u$ within $\Omega_k^{(i)}$. $F$ is a monotonically increasing imaging and post-processing operator. Then the average gray value $\mathcal{G}_k^{(i)}$ of the $k$-th class  in the $i$-th infrared image satisfies:
\begin{equation}
\label{eq:thm_average_gray_value_compact_en_final_full}
\mathcal{G}_k^{(i)}
= \mathbb{E}_{u\sim\Omega_k^{(i)}}\!\big[F(\bar L_k(u))\big]+\mathbb{E}_{u\sim\Omega_k^{(i)}}\!\big[\delta(u)\big].
\end{equation}
\end{corollary}
}

\textcolor{black}{The derivation process of Corollary \ref{corollary 1} can be viewed in Appendix. According to Eq.(\ref{eq:thm_average_gray_value_compact_en_final_full}), the gray value $\mathcal{G}_{k}^{(i)}$ of the $k$-th class is positively correlated with the effective radiation term $\mathbb{E}_{u\sim\Omega_k^{(i)}}\!\big[F(\bar L_k(u))\big]$ when the perturbation terms $\mathbb{E}_{u\sim\Omega_k^{(i)}}\!\big[\delta(u)\big]$ is within a certain small range. Thus, relative relations among class-wise gray values can reflect inter-class thermal radiation relations.}

According to the pre-annotated information of the $i$-th infrared image, we can obtain the average gray value $\mathcal{G}_{0}^{(i)}$ for the background and $\mathcal{G}_{k}^{(i)}$ for the $k$-th class, and the representation $\mathcal{D}_x$ of multi-class gray values in the scene can be expressed as:
\begin{align}
\label{eq-1}
\mathcal{D}_{x_{i}} = \{\mathcal{G}_{0}^{(i)}, \mathcal{G}_{1}^{(i)}, \mathcal{G}_{2}^{(i)},\dots, \mathcal{G}_{K}^{(i)}\}.
\end{align}

\textcolor{black}{Although the relative thermal radiation relations could be intuitively represented by the gray value ratio, e.g., $\mathcal{G}_{0}^{(i)}/\mathcal{G}_{1}^{(i)}$, this metric is inherently sensitive to the perturbation term $\mathbb{E}_{u\sim\Omega_k^{(i)}}\!\big[\delta(u)\big]$ in Eq. (\ref{eq:thm_average_gray_value_compact_en_final_full}), which is difficult to estimate under different environmental conditions. Consequently, it is necessary to develop a more robust mathematical formulation to quantify this relational characteristic.}


\textbf{Here we introduce the concept of rank}: rank is the ordinal position after the data is arranged in order, and can help us analyze the relative thermal radiation relation of the data without paying attention to the specific gray value. We can directly calculate the rank relation $\mathcal{R}(\mathcal{D}_{x_{i}})$ as:
\begin{align}
\label{eq-3}
\mathcal{R}(\mathcal{D}_{x_{i}})= Rank(\mathcal{G}_{0}^{(i)}, \mathcal{G}_{1}^{(i)}, \mathcal{G}_{2}^{(i)},\dots, \mathcal{G}_{K}^{(i)}).
\end{align}

\textcolor{black}{
Using order as a standard of measurement has \textbf{ordinal invariance:} Unlike absolute gray-value ratios, rank operation acts as a non-linear filter. For a given class pair $k$ and $\Tilde{k}$, the estimated stability remains invariant as long as the relative perturbation $\mathbb{E}_{u\sim\Omega_k^{(i)}}\!\big[\delta(u)\big]-\mathbb{E}_{u\sim\Omega_{\Tilde{k}}^{(i)}}\!\big[\delta(u)\big]$ does not exceed the intrinsic gray-value margin $\mathbb{E}_{u\sim\Omega_k^{(i)}}\!\big[F(\bar L_k(u))\big]-\mathbb{E}_{u\sim\Omega_{\Tilde{k}}^{(i)}}\!\big[F(\bar L_k(u))\big]$. This makes the metric naturally resistant to zero-mean noise and moderate adversarial attacks that do not fundamentally invert the physical hierarchy.}

After quantifying the relative thermal radiation relation for a single image $x$,  given the predicted  gray values $\mathcal{D}_{x_{i}}^{'}$ of the same image, it is necessary to define a metric to measure the discrepancy between $\mathcal{D}_{x_{i}}$ and $\mathcal{D}_{x_{i}}^{'}$. Here we apply the Spearman rank correlation coefficient to measure the differences in their thermal radiation characteristics: if the ranks between the pre-annotated result and the predicted result are exactly the same, it means that the relative gray value relation between the two results is exactly the same, that is, they have a perfect positive monotonic relation, and the Spearman rank correlation coefficient reaches the maximum value of 1; if the ranks of the two variables are completely opposite, then the coefficient is -1. Specifically, in the optimization process, the Spearman rank correlation coefficient can be mathematically formulated as follows:
\begin{align}
\label{eq-4}
\rho(\mathcal{D}_{x_{i}}^{'},\mathcal{D}_{x_{i}})=1-\frac{6\sum_{k=1}^{K}(\mathcal{R}_{k}(\mathcal{D}_{x_{i}})-\mathcal{R}_{k}(\mathcal{D}_{x_{i}}^{'}))^2}{K(K^2-1)}, 
\end{align}
where $\mathcal{R}_{k}(\mathcal{D}_{x_{i}})$ denotes the rank value of the $k$-th class in the image $x$. It should be mentioned that the calculated rank correlation coefficient is not affected by the class arrangement in $\mathcal{D}_{x_{i}}$, that is, as long as $\mathcal{D}_{x_{i}}^{'}$ and $\mathcal{D}_{x_{i}}$ are arranged in the same way, the calculated rank correlation coefficient $\rho(\mathcal{D}_{x_{i}}^{'},\mathcal{D}_{x_{i}})$ remains unchanged.

Furthermore, it is particularly emphasized that when the number of classes contained in $\mathcal{D}_{x_{i]}}$ and $\mathcal{D}_{x_{i}}^{'}$ is inconsistent, we only retain the gray values of the classes which are shared by $\mathcal{D}_{x_{i}}$ and $\mathcal{D}_{x_{i}}^{'}$, and the subsequent calculation steps are the same as those described above. 

After modeling the relative thermal radiation characteristic by $\mathcal{R}(\mathcal{D}_{x_{i}})$, we can not only measure the infrared physical knowledge, but also can further utilize $\rho(\mathcal{D}_{x_i}^{'},\mathcal{D}_{x_i})$  as an optimization goal for infrared object detection to learn the infrared physical knowledge. 

\subsection{Stability of Thermal Radiation Relation}

\textcolor{black}{
Since the thermal radiation performance is affected by environmental factors, the thermal radiation relation between different classes will change in different scenarios. Here, we try to model the stability of the thermal radiation relation. First, we model the stability of the thermal radiation relation between any two classes. Then, we extend this concept to define the stability of the overall thermal radiation relation in an image.
}

\textcolor{black}{\textbf{Stability Between Two Classes.} 
We start with the Gaussian statistical assumption in \cite{ge2020klgaussian_ir} and the region appearance modeling in \cite{rother2004grabcut}, assuming that the class-wise gray values $\mathcal{G}_k$ and $\mathcal{G}_{\Tilde{k}}$ follow Gaussian distributions:
}
\textcolor{black}{
\begin{equation}
\label{eq:gaussian_distribution_XA_XB}
\mathcal{G}_{k} \sim \mathcal{N}(\mu_{k}, \sigma_{k}^2), \qquad
\mathcal{G}_{\Tilde{k}} \sim \mathcal{N}(\mu_{\Tilde{k}}, \sigma_{\Tilde{k}}^2).
\end{equation}
}

\textcolor{black}{
According to Corollary \ref{corollary 1}, gray values serve as a digital manifestation of an object's underlying thermophysical properties. Typically, the gray value distribution within a single class is consistent, exhibiting low intra-class variance. Consequently, an obvious discrepancy in thermophysical characteristics between the $k$-th class and the $\Tilde{k}$-th class results in a pronounced separation between their respective gray value distributions. To quantify this relative relationship, we define a difference random variable $Z$ as:
}
\textcolor{black}{
\begin{equation}
\label{eq:Z_(valuable)}
Z = \mathcal{G}_{k} - \mathcal{G}_{\Tilde{k}},
\end{equation}
}
\textcolor{black}{
where the sign of $Z$ reflects the relative gray value relationship between classes $k$ and $\Tilde{k}$, and the  probability $\mathbb{P}(Z > 0)$ or $\mathbb{P}(Z < 0)$ approaching 1 indicates a consistent gray value order and thus a stable relative thermal radiation relation. Thus, the stability of the relative thermal radiation relation can be defined by the absolute gap between $\mathbb{P}(Z > 0)$ and $\mathbb{P}(Z < 0)$, combined with Eq.(\ref{eq:gaussian_distribution_XA_XB}), we can obtain the stability $\phi_{k\Tilde{k}}$ between classes $k$ and $\Tilde{k}$ as follows:
}
\textcolor{black}{
\begin{align}
\phi_{k\Tilde{k}} = \left| 1 - 2\Phi\left( -\frac{\mu_{k} - \mu_{\Tilde{k}}}{\sqrt{\sigma_{k}^2 + \sigma_{\Tilde{k}}^2}} \right) \right|,
\end{align}}
\textcolor{black}{and the detailed derivation process can be viewed in Appendix. The closer $\phi_{k\Tilde{k}}$ is to 1, the more pronounced the separation between the gray value distributions of the two classes, signifying a more stable relative thermal radiation relation. Conversely, as $\phi_{k\Tilde{k}}$ approaches 0, the distributions overlap more significantly, indicating a more volatile and less certain relationship.
}

\textcolor{black}{
However, as the underlying distribution parameters (such as mean $\mu_{k}$ and variance $\sigma_{k}$) are typically unobservable, we resort to an empirical estimation through sampling. Let $N_{k\Tilde{k}}$ denote the number of images where classes $k$-th and $\Tilde{k}$-th co-occur. Given a sufficiently large $N_{k\Tilde{k}}$ , the stability $\phi_{k\Tilde{k}}$ can be approximated as follows:
}
\textcolor{black}{
\begin{equation}
\label{eq:phi_varphi_compact}
\phi_{k\Tilde{k}} \approx \varphi_{k\Tilde{k}} \triangleq \left| \frac{1}{N_{k\Tilde{k}}} \sum_{i=1}^{N_{k\Tilde{k}}} \operatorname{sgn} \left( \mathcal{R}_k(\mathcal{D}_{x_i}) - \mathcal{R}_{\Tilde{k}}(\mathcal{D}_{x_i}) \right) \right|,
\end{equation}
}
\textcolor{black}{where $\mathcal{R}_k(\mathcal{D}_{x_i})$ and $\mathcal{R}_{\Tilde{k}}(\mathcal{D}_{x_i})$ represent the rank positions of the $k$-th and $\Tilde{k}$-th classes in the $i$-th image, respectively. $\operatorname{sgn}(\cdot)$ is the sign function that returns $1$ if the value inside the parentheses is greater than 0, $-1$ if it is less than 0, and $0$ if it is equal to 0. The detailed derivations can be found in the Appendix.
}




\textcolor{black}{\textbf{Stability in a Single Image.}}
For the image $x_i$, we assume that it contains $K_i$ distinct classes. When the relative thermal radiation relations among these classes remain stable, the overall relative thermal radiation relation $\mathcal{R}(\mathcal{D}_{x_i})$ can be considered stable; conversely, if the relative thermal radiation relations among the $K_i$ classes exhibit obvious fluctuations, the $\mathcal{R}(\mathcal{D}_{x_i})$ is accordingly regarded as unstable. 

\begin{figure*}[t]
    \centering
    \includegraphics[width=1\textwidth]{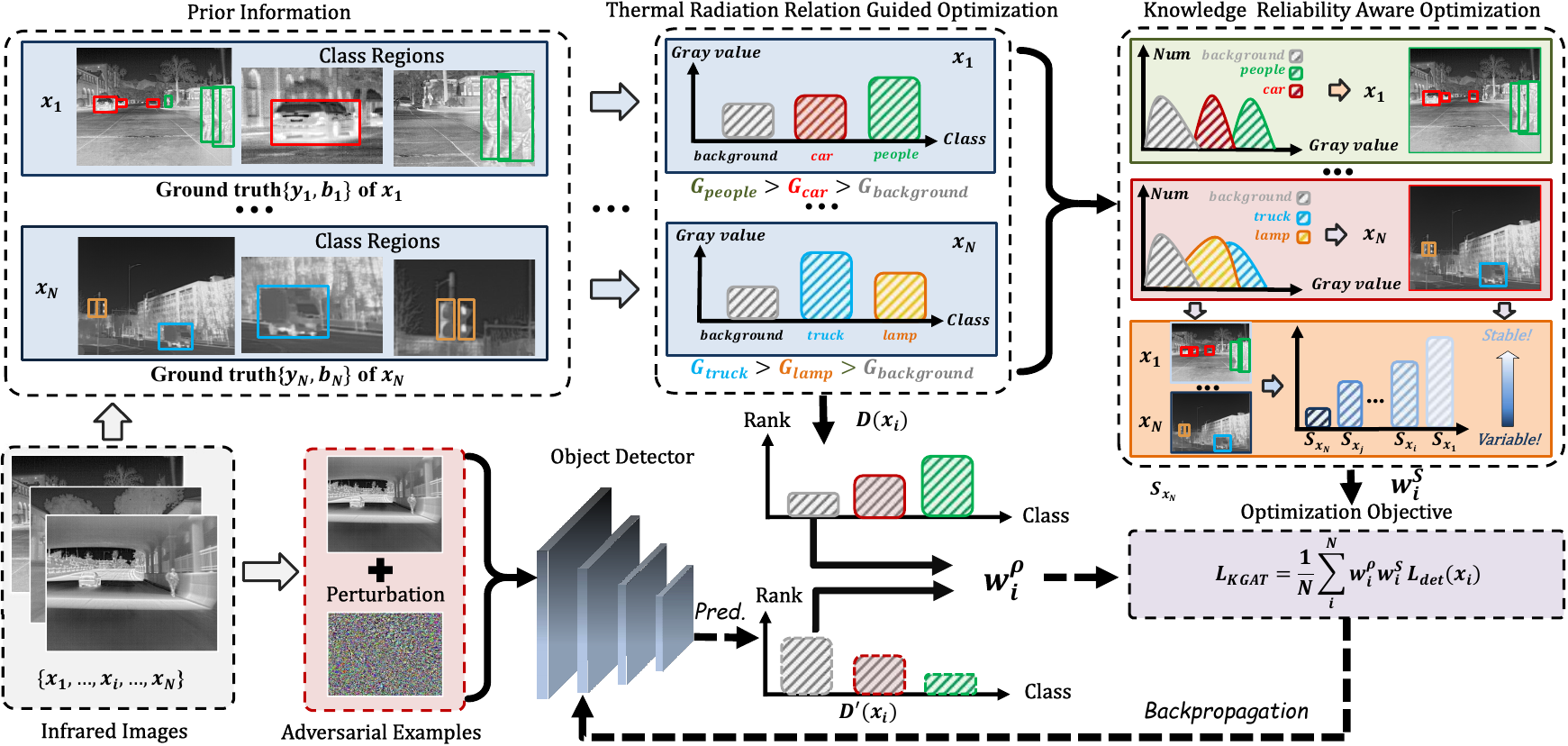}
    \caption{\textbf{The Framework of our Knowledge-Guided Adversarial Training.} \textbf{During the knowledge extraction process},  we first calculate the rank order of the gray value between different classes to extract the thermal radiation relation. We further calculate the stability of the thermal radiation relation between all the different classes. \textbf{During the training process}, we apply the thermal radiation relations and the corresponding variations to adjust the learning weights of each training sample so that the prediction results conform to the real-world infrared knowledge.}
    \label{fig:framework}
\end{figure*}

Thus, the stability of pairwise relative thermal radiation relations among these $K_i$ classes can be used to quantify the stability of the overall thermal radiation relation $\mathcal{R}(\mathcal{D}_{x_i})$ in the image $x_i$, which can be represented as follows:
\begin{align}
\label{delta}
\mathcal{S}_{x_i} =  \frac{1}{K_i(K_i-1)}\sum\limits_{\Tilde{k}_{i}=1,\Tilde{k}_{i}\neq k_{i}}^{K_i} \varphi_{k_{i}\Tilde{k}_{i}}.
\end{align}

When the relative thermal radiation relation between any class $k_i$ and other class $\Tilde{k}_i$ tends to be stable (i.e., $\varphi_{k_i\Tilde{k}_i}$ approaches 1), $\mathcal{S}_{x_i}$ approaches 1, indicating that the overall thermal radiation relation $\mathcal{R}(\mathcal{D}_{x_i})$ is stable. Conversely, when $\mathcal{S}_{x_i}$ approaches 0, it indicates that the overall relative thermal radiation relation tends to be volatile.

Thus, based on the above analysis, $\mathcal{S}_{x_{i}}$ can accurately quantify the stability of the relative thermal radiation relations exhibited in the current infrared scene, which can further guide the infrared object detector to allocate more learning focus to the stable patterns of relative thermal radiation relations that inherently reflect the infrared physical property differences between classes and are more robust to various interfering factors.

\section{Knowledge-guided Adversarial Training}
\label{sec:method}

\subsection{Overall Framework}
Building upon the theoretical underpinnings established in Section 3, 
we hypothesize that adversarial robustness can be significantly bolstered by embedding the relative thermal radiation relation, which serves as a form of infrared physical knowledge, in the adversarial training process of infrared images. Therefore, based on the adversarial training framework \citep{zhang2019mtd,Dong2022AdversariallyAware}, we propose a knowledge-guided adversarial training method to guide the predicted result to match the relative thermal radiation relation. And our optimization goal can be formulated as follows:
\begin{align}
\label{sec4:eq-1}
    \mathop{\arg\min}_{\theta}\mathbb{E}_{(x,\{y,b\})\sim \mathcal{C}}\mathcal{L}_{KGAT}(f_{\theta}(\Tilde{x}),\{y,b\}), 
\end{align}
where $\mathcal{L}_{KGAT}$ denotes the optimization loss function of KGAT, containing classification and localization loss guided by the infrared thermal radiation relation, and $\{y,b\}$ denotes the ground truth including class label $y$ and the bounding box $b$. As for adversarial examples $\Tilde{x}$, we follow the MTD setting in \cite{zhang2019mtd} and the SALC setting in \cite{Chen2024Accurate},  where one selects the attack type with the highest attack performance from the classification and localization task domains, and the other selects the attack type with the highest performance from the multi-task combination domains. 

\subsection{Thermal Radiation Relation Guided Optimization}
\label{Thermal Radiation Relation Guided Optimization}

Here, we aim to impose constraints of thermal radiation relations on model prediction results, enabling the model to fully learn the potential physical knowledge in the infrared images and make predictions consistent with the actual physical world, finally improving the performance of the infrared object detection. From the intuitive point of view, the loss function of infrared object detection guided by knowledge can be changed into the following form:
\begin{align}
\label{eq:goal}
{\mathcal{L}^{'}}_{KGAT}=\mathcal{L}_{det}+\alpha\mathcal{L}_{knowledge},
\end{align}
where $\mathcal{L}_{det}$ and $\mathcal{L}_{knowledge}$ represent the loss function of object detection and infrared knowledge, respectively.  with $\alpha$ serving as the important trade-off hyperparameter.

Based on the above analysis, we pre-count the gray value relation $\mathcal{D}_{x}$ between the classes in the infrared images. As for the prediction results of the infrared object detection, we can roughly calculate the gray value relation $\mathcal{D}_{x}^{'}$ between different infrared classes through the bounding box predicted by the infrared object detection. Then, according to the theoretical modeling in the above subsection, we can measure the gap between $\mathcal{D}_{x}$ and $\mathcal{D}_{x}^{'}$ by the Spearman rank correlation $\rho$: the larger $\rho$ means the more similar. Here we can define the infrared knowledge loss $\mathcal{L}_{knowledge}$ as follows:
\begin{align}
\label{eq:L_knowledge}
\mathcal{L}_{knowledge}=1-\rho(\mathcal{D}_{x}^{'},\mathcal{D}_{x}).
\end{align}

However, the infrared knowledge loss $\mathcal{L}_{knowledge}$  is non-differentiable due to its reliance on discrete relational constraints and discontinuous transformations. This property precludes the direct application of gradient-based optimization methods such as backpropagation. Fortunately, empirical observations suggest a monotonic alignment between the detection loss $\mathcal{L}_{det}$ and infrared knowledge loss $\mathcal{L}_{knowledge}$. Specifically, a low $\mathcal{L}_{det}$ implies highly accurate bounding box predictions, which in turn corresponds to a low $\mathcal{L}_{knowledge}$. 

Therefore, rather than directly minimizing $\mathcal{L}_{knowledge}$ , we propose an indirect optimization strategy,  which is formulated based on Eq. (8) as follows: (\ref{eq:L_knowledge}), which can be formulated as follows:

\begin{align}
\label{eq:loss}
\mathcal{L}_{KGAT}^{'} = \frac{1}{N}\sum_i^Nw_i^{\rho}\mathcal{L}^{i}_{det},
\end{align}
where $\mathcal{L}_{det}^{i}$ denotes the detection loss for $i$-th image $x_i$.  The optimization function converts the Spearman rank correlation coefficient $\rho$ into the weights of $\mathcal{L}_{det}$ for different images. If the infrared knowledge loss $\mathcal{L}_{knowledge}$ is not well optimized, it is necessary to increase the optimization strength $w_i^{\rho}$ of the corresponding image $x_i$ to increase the corresponding $\mathcal{L}_{det}$, further learn the thermal radiation relation and reduce the infrared knowledge loss $\mathcal{L}_{knowledge}$. Here we provide Theorem  \ref{thorem 1} to demonstrate the advantage of the loss $\mathcal{L}'_{KGAT}$ in minimizing the error risk expectation of infrared knowledge loss $\mathcal{L}_{knowledge}$.

\begin{Theorem}
\label{thorem 1}
In an infrared scene dataset $\mathcal{C}$, the initial parameter of infrared object detector $M_{\theta}$ is $\theta$.  $\theta_1$ denotes the new parameter updated with the loss function $\mathcal{L}_{det}$, while $\theta_2$ denotes the new parameter updated with the  $\mathcal{L}'_{KGAT}$. The error risk expectation $E_{\theta_1}[\mathcal{L}_{knowledge}]$ of $\theta_1$ and error risk expectation $E_{\theta_2}[\mathcal{L}_{knowledge}]$ of $\theta_2$ have the relationship as follows:
\begin{align}
\label{expectation}
E_{\theta_1}[\mathcal{L}_{knowledge}]>E_{\theta_2}[\mathcal{L}_{knowledge}].
\end{align}
\end{Theorem}

The proof of Theorem \ref{thorem 1} can be viewed in Appendix. Theorem \ref{thorem 1} demonstrates that, compared with directly using $\mathcal{L}_{det}$ as the loss function, our indirect optimization idea can more effectively reduce the expected loss of $\mathcal{L}_{knowledge}$ over the entire dataset, which further demonstrates the necessity and effectiveness of converting $\mathcal{L}_{knowledge}$ loss into the weights of the $\mathcal{L}_{det}$ loss in our method.

\begin{algorithm}[t]  
  \caption{Knowledge-Guided Adversarial Training}  
  \label{algorithm:1}
  \begin{algorithmic}[1]   
   \Require {The train dataset $\mathcal{D}$, clean examples $x$, the ground truth $\{y,b\}$ including class label $y$ and the bounding box $b$, infrared object detection $f_{\theta}$ with model parameter $\theta$, total class number $C$, the max training epochs $max$-$epoch$, batch size $N$, weight learning rate $\eta$.}         
    \State {  \small Get thermal radiation relation $\mathcal{R}(\mathcal{D}_{x})$ based on Eq.(\ref{eq-3}).}    
    \State {  \small  Get thermal radiation relation variation $\varphi_{k\Tilde{k}}$ between class $k$ and $\Tilde{k}$ based on Eq.(\ref{eq:phi_varphi_compact}).} 
     \For{$0$ to $max$-$epoch$} 
        \For{$Every~minibatch(x,t,y)~in~\mathcal{D}$}
             \State {  \small$\Tilde{x}=\mathop{argmax}\limits_{||\Tilde{x}-x|| \leq \epsilon} \mathcal{L}_{det}(f_{\theta}(\Tilde{x}),\{y,b\})$.} 
             \State {  \small$\mathcal{R}(\mathcal{D}_{\Tilde{x}})= Rank(\mathcal{G}_{0}, \mathcal{G}_{1}, \mathcal{G}_{2},\dots, \mathcal{G}_{K})$.}      
            \For{$each ~\Tilde{x}_i~ in~\Tilde{x}$}            
          \State {\small  $w_i^{\rho} = -log(\beta\cdot\rho(\mathcal{D}_{\Tilde{x}_i}^{'},\mathcal{D}_{x_i}) + 1) +1$.}  
          \State {\small  $w_i^{\mathcal{S}} = (\upsilon + \eta\mathcal{S}_{x_i})^\gamma$.}  
          \EndFor  
        \State { \small $\theta = \theta - \eta \cdot \nabla_{\theta} \frac{1}{N}\sum_i^Nw_i^{\rho}w_i^{\mathcal{S}}\mathcal{L}^{i}_{det}$.}   
          \EndFor  
    \EndFor  
  \end{algorithmic}  
\end{algorithm}

Based on the above analysis, during the optimization process, we design a sample weight $w_i^{\rho}$ for the $i$-th image $x_i$ based on the current correlation coefficient, and the sample weight $w_i^{\rho}$ is negatively correlated with the correlation coefficient, which can be formulated as follows:
\begin{align}
\label{eq:weight}
w_i^{\rho} = -log(\beta\cdot\rho(\mathcal{D}_{x_i}^{'},\mathcal{D}_{x_i}) + 1) +1, \beta \in (0, 1),
\end{align}
where $\beta$ is a hyper-parameter to control the adjustment strength, a larger value of $\beta$ causes $w_i^{\rho}$ to increase more rapidly as $\rho$ decreases. \textcolor{black}{The weight $w_i^{\rho}$ prioritizes the learning of poorly predicted relations (low $\rho_i$) by assigning them larger optimization weights, while moderately down-weighting those already well-predicted  (high $\rho_i$). This nonlinear reweighting mechanism is characterized by its smoothness and boundedness, which effectively prevents weight explosion and ensures numerical stability during training. The formal derivation is provided in the Appendix.}
A key point is that although the weight $w_i^{\rho}$ is a function of $\rho(\mathcal{D}_{x}^{'},\mathcal{D}_{x})$, it is treated as a optimization constant.


In addition, since we use the model's prediction box as the basis for calculating the relative gray value relation, and due to the existence of the NMS operation, many candidate boxes will be filtered out, so directly applying the model's final prediction box from object detection will not fully utilize the model's prediction information to a large extent. Therefore, we select a candidate box setting with high confidence (we set it to 0.05) for this class before the NMS operation to calculate the gray value relation $\mathcal{D}_{x}^{'}$ as the final version and further obtain the final Spearman rank correlation coefficient $\rho$. For DETR-based methods, such as \cite{carion2020detr,zhu2021deformabledetr,xie2021maskdino,zhang2022dino}, since they do not rely on the NMS operation, we directly use the prediction result of the query that best matches the label.

\subsection{Knowledge Reliability Aware Optimization}
\label{Knowledge Reliability Aware Optimization}

In addition,  due to the interference of complex external factors such as environmental conditions, the stability of the thermal radiation relation varies across the distinct scenes presented by different images. Therefore, from the perspective of optimization, to better utilize those stable thermal radiation relations during target recognition, for those images which contain such stable thermal radiation relations, we need to increase the optimization strength on the corresponding examples, while for images containing thermal radiation relation with obvious fluctuations, we can appropriately reduce the optimization strength, so that the infrared object detector can acquire more robust infrared thermal radiation relation that serves as a form of infrared physical knowledge and is generalizable across scenarios.

Therefore, we propose the knowledge reliability aware optimization that is aware of the reliability of the thermal radiation relation to enhance the effectiveness of adversarial training. First, for image $x_i$, we quantify the stability $\mathcal{S}_{x_i}$ of the infrared thermal radiation relation within it based on Eq. (\ref{delta}), which reflects the reliability of the infrared thermal radiation relation contained $x_i$; subsequently, based on $\mathcal{S}_{x_i}$, we calculate the optimization strength  to be applied to the image $x_i$, which can be formulated as follows:
\begin{align}
\label{eq:weight_1}
w_i^{\mathcal{S}} = (\upsilon + \eta\mathcal{S}_{x_i})^\gamma,
\end{align}
where $\upsilon$ and $\eta$ normalize the weights into the value range around 1 (we set the $\upsilon$ to 0.95 and $\eta$ to 0.1), and the hyper-parameter $\gamma$ is applied as a control parameter of optimization strength. Then we can combine it with thermal radiation relation guided optimization, and the final optimization goal of KGAT can be formulated as follows:
\begin{align}
\label{eq:final loss}
\mathcal{L}_{KGAT} &=\frac{1}{N}\sum_i^Nw_i^{\rho}w_i^{\mathcal{S}}\mathcal{L}^{i}_{det}.
\end{align}

The entire framework of KGAT can be viewed in Figure \ref{fig:framework}, and the algorithm can be viewed in Figure \ref{algorithm:1}. In summary, we embed thermal radiation relation into the adversarial training process of infrared object detectors by adjusting the weights of the optimization loss, which realizes the dual-wheel drive of data and knowledge. Our Knowledge-Guided adversarial training can effectively make the model's prediction results more consistent with the cognitive real laws of thermal radiation in the physical world and can potentially improve the robustness towards the diverse adversarial attacks and common corruptions.

\begin{table*}[t]
\renewcommand{\arraystretch}{1.3}
\setlength{\tabcolsep}{0pt}
\centering
\caption{Robustness performance of different models under  various adversarial attacks and common corruptions on M$^3$FD. All results are based on the best checkpoints. \textbf{GN} denotes Gaussian Noise, and \textbf{Avg} denotes the average value of the overall mAP$_{50}$. $\textcolor{darkgreen}{\downarrow}$ and $\textcolor{red}{\uparrow}$ indicate the mAP$_{50}$ decrease or increase for the current version of the KGAT method compared to baseline, respectively.}
\label{main_experiment_M3FD}
\scalebox{0.68}{
\begin{tabular}{c|c|ccccccccc>{\columncolor{gray!10}}c}
\noalign{\vskip 0mm \hrule height 0.5mm \vskip 0mm}
Model        & Method               & Clean$\phantom{{\downarrow}{18.6}}$  & A$_{cls}$$\phantom{{\downarrow}{18.6}}$ & A$_{loc}$$\phantom{{\downarrow}{18.6}}$ & A$_{mtd}$$\phantom{{\downarrow}{18.6}}$ & CWA$\phantom{{\downarrow}{18.6}}$   & DAG$\phantom{{\downarrow}{18.6}}$   & Blur$\phantom{{\downarrow}{18.6}}$  & Salt$\phantom{{\downarrow}{18.6}}$  & GN$\phantom{{\downarrow}{18.6}}$ & \textbf{Avg}$\phantom{{\downarrow}{18.6}}$ \\
\hline
\multirow{11}{*}{SSD} & STD        & ${\underline{71.3}\phantom{{\downarrow}{18.6}}}$ & ${9.8}\phantom{{\downarrow}{18.6}}$    & ${9.2}\phantom{{\downarrow}{18.6}}$    & ${8.9}\phantom{{\downarrow}{18.6}}$  &  ${8.4}\phantom{{\downarrow}{18.6}}$ & ${10.5}\phantom{{\downarrow}{18.6}}$  & ${47.6}\phantom{{\downarrow}{18.6}}$ & ${50.1}\phantom{{\downarrow}{18.6}}$ & ${40.9}\phantom{{\downarrow}{18.6}}$ &
${24.9}\phantom{{\downarrow}{18.6}}$\\
                     & LOC \citep{zhang2019mtd}        & ${47.9}\phantom{{\downarrow}{18.6}}$ & ${29.3}\phantom{{\downarrow}{18.6}}$   & ${29.7}\phantom{{\downarrow}{18.6}}$   & ${29.0}\phantom{{\downarrow}{18.6}}$ & 
                     ${26.7}\phantom{{\downarrow}{18.6}}$ & 
                     ${38.3}\phantom{{\downarrow}{18.6}}$ & ${33.3}\phantom{{\downarrow}{18.6}}$ & ${36.8}\phantom{{\downarrow}{18.6}}$ & ${31.1}\phantom{{\downarrow}{18.6}}$ 
                     & ${33.6}\phantom{{\downarrow}{18.6}}$\\
\end{tabular}
}
\end{table*}
\begin{table*}[t]
\renewcommand{\arraystretch}{1.3}
\setlength{\tabcolsep}{0.1pt}
\centering
\caption{Robustness performance of different models under  various adversarial attacks and common corruptions on FLIR-ADAS. All results are based on the best checkpoints. \textbf{GN} denotes Gaussian Noise, and \textbf{Avg} denotes the average value of the overall mAP$_{50}$. $\textcolor{darkgreen}{\downarrow}$ and $\textcolor{red}{\uparrow}$ indicate the mAP$_{50}$ decrease or increase for the current version of the KGAT method compared to baseline, respectively.} 
\label{main_experiment_flir_adav}
\scalebox{0.68}{
\begin{tabular}{c|c|ccccccccc>{\columncolor{gray!10}}c}
\noalign{\vskip 0mm \hrule height 0.5mm \vskip 0mm}
Model        & Method                & Clean$\phantom{{\downarrow}{18.6}}$  & A$_{cls}$$\phantom{{\downarrow}{18.6}}$ & A$_{loc}$$\phantom{{\downarrow}{18.6}}$ & A$_{mtd}$$\phantom{{\downarrow}{18.6}}$ & CWA$\phantom{{\downarrow}{18.6}}$   & DAG$\phantom{{\downarrow}{18.6}}$   & Blur$\phantom{{\downarrow}{18.6}}$  & Salt$\phantom{{\downarrow}{18.6}}$  & GN$\phantom{{\downarrow}{18.6}}$ &
\textbf{Avg}$\phantom{{\downarrow}{18.6}}$\\
\hline
\multirow{11}{*}{SSD} & \text{STD} & $\underline{61.9}\phantom{{\downarrow}{18.6}}$ & ${8.8}\phantom{{\downarrow}{18.6}}$ & ${8.5}\phantom{{\downarrow}{18.6}}$ & ${7.9}\phantom{{\downarrow}{18.6}}$ &
${5.6}\phantom{{\downarrow}{18.6}}$ & 
${10.3}\phantom{{\downarrow}{18.6}}$ & ${44.1}\phantom{{\downarrow}{18.6}}$ & ${47.6}\phantom{{\downarrow}{18.6}}$ & ${38.9}\phantom{{\downarrow}{18.6}}$ &
${22.7}\phantom{{\downarrow}{18.6}}$\\
                        & LOC \citep{zhang2019mtd}&${36.6}\phantom{{\downarrow}{18.6}}$ & ${26.6}\phantom{{\downarrow}{18.6}}$ & ${27.2}\phantom{{\downarrow}{18.6}}$ & ${26.9}\phantom{{\downarrow}{18.6}}$ & 
                        ${21.5}\phantom{{\downarrow}{18.6}}$ & 
                        ${29.2}\phantom{{\downarrow}{18.6}}$ & ${30.6}\phantom{{\downarrow}{18.6}}$ & ${34.4}\phantom{{\downarrow}{18.6}}$ & ${28.5}\phantom{{\downarrow}{18.6}}$& ${29.1}\phantom{{\downarrow}{18.6}}$ \\
                        & CLS \citep{zhang2019mtd} & ${38.1}\phantom{{\downarrow}{18.6}}$ & ${28.3}\phantom{{\downarrow}{18.6}}$ & ${28.7}\phantom{{\downarrow}{18.6}}$ & 
                        ${28.4}\phantom{{\downarrow}{18.6}}$ & 
                        ${23.2}\phantom{{\downarrow}{18.6}}$ & 
                        ${30.6}\phantom{{\downarrow}{18.6}}$ & ${33.1}\phantom{{\downarrow}{18.6}}$ & ${35.4}\phantom{{\downarrow}{18.6}}$ & ${30.7}\phantom{{\downarrow}{18.6}}$ & ${30.8}\phantom{{\downarrow}{18.6}}$\\
\end{tabular}
}
\end{table*}

\begin{table*}[t]
\renewcommand{\arraystretch}{1.3}
\setlength{\tabcolsep}{0.1pt}
\centering
\caption{Robustness performance of different models under  various adversarial attacks and common corruptions on KAIST. All results are based on the best checkpoints. \textbf{GN} denotes Gaussian Noise, and \textbf{Avg} denotes the average value of the overall mAP$_{50}$. $\textcolor{darkgreen}{\downarrow}$ and $\textcolor{red}{\uparrow}$ indicate the mAP$_{50}$ decrease or increase for the current version of the KGAT method compared to baseline, respectively.} 
\label{main_experiment_KAIST}
\scalebox{0.68}{
\begin{tabular}{c|c|ccccccccc>{\columncolor{gray!10}}c}
\noalign{\vskip 0mm \hrule height 0.5mm \vskip 0mm}
Model        & Method                & Clean$\phantom{{\downarrow}{18.6}}$  & A$_{cls}$$\phantom{{\downarrow}{18.6}}$ & A$_{loc}$$\phantom{{\downarrow}{18.6}}$ & A$_{mtd}$$\phantom{{\downarrow}{18.6}}$   & 
CWA$\phantom{{\downarrow}{18.6}}$   &
DAG$\phantom{{\downarrow}{18.6}}$   & Blur$\phantom{{\downarrow}{18.6}}$  & Salt$\phantom{{\downarrow}{18.6}}$  & GN$\phantom{{\downarrow}{18.6}}$  &
\textbf{Avg}$\phantom{{\downarrow}{18.6}}$\\
\hline
\multirow{11}{*}{SSD}       & STD                  & $\underline{62.2}\phantom{{\downarrow}{18.6}}$  & ${8.4}\phantom{{\downarrow}{18.6}}$    & ${8.2}\phantom{{\downarrow}{18.6}}$    & ${8.1}\phantom{{\downarrow}{18.6}}$   & 
${7.7}\phantom{{\downarrow}{18.6}}$   & 
${7.9}\phantom{{\downarrow}{18.6}}$   & $\textbf{43.5}\phantom{{\downarrow}{18.6}}$  & $\underline{49.5}\phantom{{\downarrow}{18.6}}$  & ${41.1}\phantom{{\downarrow}{18.6}}$ &
${24.1}\phantom{{\downarrow}{18.6}}$\\
             & LOC \citep{zhang2019mtd}                 & ${39.9}\phantom{{\downarrow}{18.6}}$  & ${28.3}\phantom{{\downarrow}{18.6}}$   & ${29.8}\phantom{{\downarrow}{18.6}}$   & ${28.0}\phantom{{\downarrow}{18.6}}$  & 
             ${27.5}\phantom{{\downarrow}{18.6}}$ & 
             ${33.0}\phantom{{\downarrow}{18.6}}$  & ${29.2}\phantom{{\downarrow}{18.6}}$  & ${30.1}\phantom{{\downarrow}{18.6}}$  & ${27.3}\phantom{{\downarrow}{18.6}}$ &
             ${30.3}\phantom{{\downarrow}{18.6}}$\\
\end{tabular}
}
\end{table*}
\section{Experiment}
\label{sec:experiment}

\subsection{Experimental Settings}

\noindent \textbf{Datasets.} To verify the generalization performance across different infrared imaging devices, we evaluate  on three multi-class infrared datasets for autonomous driving and pedestrian detection scenarios: M$^3$FD \citep{liu2022target}, FLIR-ADAS \citep{flir_adas}, and KAIST \citep{hwang2015multispectral}. 
The M$^3$FD dataset is an infrared and RGB object detection dataset containing 4,200 images, with annotations for  6 classes, including people, cars, buses, motorcycles, trucks, and lights. 
The images in the FLIR-ADAS dataset include infrared bands, which have the characteristics of excellent performance in low-light and complex background environments. The dataset includes 15 classes and more than 9,700 infrared images with object annotations. 
KAIST contains 95,328 aligned color-thermal frame pairs, with 103,128 dense annotations on 1,182 unique pedestrians. Due to the limited computational resources and the need to maximize the diversity of data, we select the training and test sub-sets of KAIST following \cite{liu2016multispectral}. Ultimately, we obtain 7,601 training images and 2,252 test images.

\noindent \textbf{Model Selections.} We conduct the experiments with four different types of object detection models to verify the effectiveness and generalization of the KGAT method: YOLO-v8 \citep{Jocher_Ultralytics_YOLO_2023}, Faster R-CNN \citep{ren2016faster}, DINO \citep{zhang2022dino}, and SSD \citep{liu2016ssd}. YOLO-v8 and SSD are single-stage object detectors, while Faster R-CNN is a two-stage object detector. Additionally, DINO, unlike these models, is a single-stage object detector based on the transformer architecture.

\noindent \textbf{Baselines.} Here we select eight different methods for comparison: STD, LOC \citep{zhang2019mtd}, CLS \citep{zhang2019mtd}, MTD \citep{zhang2019mtd}, CWAT \citep{chen2021class}, Det-AdvProp \citep{chen2021robust}, SALC \citep{Chen2024Accurate}, and RobustNet \citep{Dong2022AdversariallyAware}. STD indicates that the model is trained on clean images using the standard training method. Meanwhile, LOC, CLS, MTD, and CWAT are pure adversarial training methods; Det-AdvProp, SALC, and RobustNet combine architectural modifications with adversarial training. 
It should be mentioned that the dynamic convolution technology on RobustDet will bring a large computational overhead when transferred to other models, so we only conducted relevant experiments on SSD following its original paper.

\noindent \textbf{Evaluation Metric.} Following \cite{zhang2019mtd,Chen2024Accurate,Dong2022AdversariallyAware}, we use the mainstream metric mAP$_{50}$ to assess the performance on the object detection task. To fully evaluate the robustness, besides the clean accuracy, we mainly include 9 different adversarial attacks and common corruptions, where adversarial attacks include three PGD-based Attacks \citep{madry2018pgd} (Classification Attack $\boldsymbol{\text{A}_{cls}}$, Localization Attack $\boldsymbol{\text{A}_{loc}}$, MTD Attack $\boldsymbol{\text{A}_{mtd}}$) in \cite{zhang2019mtd},  CWA Attack \citep{chen2021class}, DAG Attack \citep{xie2017adversarial}, Adversarial Infrared Patches (AIP) attack \citep{wei2023physically}, common corruptions include Pixel blurring, Gaussian noise, and Salt noise. 

For A$_{cls}$ and A$_{loc}$ attacks, we follow \cite{Dong2022AdversariallyAware} and utilize the classification and localization branches to generate adversarial examples, respectively. For the A$_{mtd}$ attack, we follow the setting in \cite{zhang2019mtd} to alternate use classification and localization branches.  Simultaneously, the $\ell_\infty$ norm is employed to quantify the magnitude of adversarial perturbations. Note that all PGD attacks use 10 iterations and a maximum perturbation of 8/255. For the CWA attack, we follow the setting in \cite{chen2021class}. For the DAG attack, we follow the setup in \cite{xie2017adversarial} and perform 150 steps to make an effective attack. For Pixel blurring, we apply a Gaussian blur with a radius of 4 pixels. For salt noise, we utilize 0.01 as the mutation probability for each pixel. For Gaussian noise, we sample the noise from a Gaussian distribution $\mathcal{N}(0, \sigma^2)$, where $\sigma$ is set to 6, to simulate the strong and random noise interference that may occur during thermal infrared imaging.

\begin{figure*}[t]
    \centering
\includegraphics[width=0.9\linewidth]{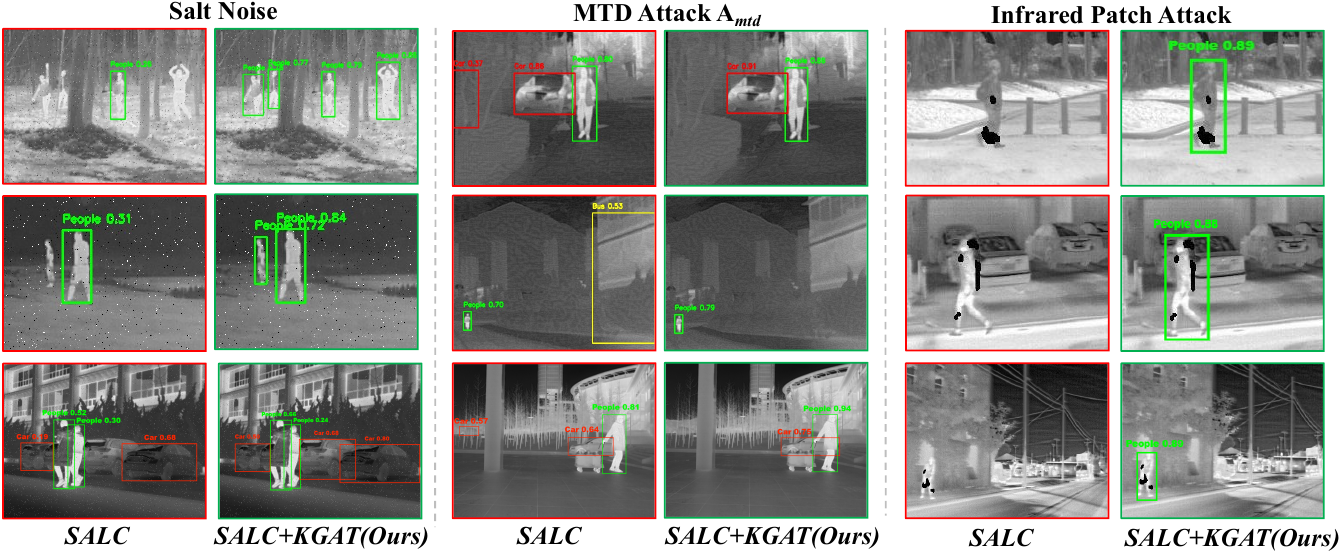}
    \caption{Visualization of the prediction results between baseline (SALC) and our SALC+KGAT of YOLO-v8 on M$^3$FD under three different attacks. 
} \label{fig:image Amtd}
\end{figure*}

\noindent \textbf{Training Setting.} 
For all the infrared object detections, we train for a total of 100 epochs with a batch size of 16. For YOLO-v8, we use YOLO-v8-x, and the weight optimizer is based on SGD, while the initial rate is 0.01 with momentum 0.937, and weight decay is $5\times10^{-4}$ following the original setting. For Faster R-CNN, we use ResNet-101 as the backbone, the weight optimizer is SGD, the initial rate is 0.01 with momentum 0.9, and weight decay is $1\times10^{-4}$.
For SSD, we use SSD300 with VGG-16 as the backbone, the weight optimizer is SGD, the initial rate is 0.001 with momentum 0.9, and weight decay is $5\times10^{-4}$. For DINO, we use InternImage as the backbone, the weight optimizer is Adam, and the initial rate is set to $1\times10^{-4}$ with momentum 0.9. As for the important hyperparameters $\beta$ and $\gamma$, we set them to 0.5 and 1.1 without additional instructions.

Due to the substantial GPU requirements in the training phase, all experiments, except for those involving DINO, were conducted on the RTX 4090, whereas those for DINO were trained on the A100-80G. Checkpoints for evaluation are selected based on the sum of clean accuracy and adversarial robustness against MTD attack.

\begin{table*}[t]
\renewcommand{\arraystretch}{1.3}
\setlength{\tabcolsep}{0.1pt}
\centering
\caption{\textcolor{black}{Robustness performance of Specialized infrared detection architectures under  various adversarial attacks and common corruptions on M$^3$FD. All results are based on the best checkpoints. \textbf{GN} denotes Gaussian Noise, and \textbf{Avg} denotes the average value of the overall mAP$_{50}$. }} 
\label{detection_result}
\scalebox{0.9}{
\begin{tabular}{c|c|ccccccccc>{\columncolor{gray!10}}c}
\noalign{\vskip 0mm \hrule height 0.5mm \vskip 0mm}
Model        & Method                & \phantom{0}Clean$\phantom{{\downarrow}{1 }}$  & A$_{cls}$$\phantom{{\downarrow}{1 }}$ & A$_{loc}$$\phantom{{\downarrow}{1 }}$ & A$_{mtd}$$\phantom{{\downarrow}{1 }}$ & CWA$\phantom{{\downarrow}{1 }}$   & DAG$\phantom{{\downarrow}{1 }}$   & Blur$\phantom{{\downarrow}{1 }}$  & Salt$\phantom{{\downarrow}{1 }}$  & GN$\phantom{{\downarrow}{1 }}$ &
\textbf{Avg}$\phantom{{\downarrow}{1 }}$\\
\hline
\multirow{3}{*}{EFLNet}\phantom{0} & \text{STD} & \phantom{0}$\textbf{88.1} $$\phantom{{\downarrow}{1 }}$ & ${8.9} $$\phantom{{\downarrow}{1 }}$ & ${7.3} $$\phantom{{\downarrow}{1 }}$ & ${6.4} $$\phantom{{\downarrow}{1 }}$ &
${12.7} $$\phantom{{\downarrow}{1 }}$ & 
${2.1} $$\phantom{{\downarrow}{1 }}$ & ${33.0} $$\phantom{{\downarrow}{1 }}$ & ${53.5} $$\phantom{{\downarrow}{1 }}$ & $\textbf{65.9} $$\phantom{{\downarrow}{1 }}$ &
${30.9} $$\phantom{{\downarrow}{1 }}$\\
 &\text{SALC} & \phantom{0}${80.2} $$\phantom{{\downarrow}{1 }}$ & ${65.2} $$\phantom{{\downarrow}{1 }}$ & ${66.8} $$\phantom{{\downarrow}{1 }}$ & ${63.6} $$\phantom{{\downarrow}{1 }}$ &
${60.3} $$\phantom{{\downarrow}{1 }}$ & 
${61.6} $$\phantom{{\downarrow}{1 }}$ & ${46.3} $$\phantom{{\downarrow}{1 }}$ & ${54.3} $$\phantom{{\downarrow}{1 }}$ & ${64.3} $$\phantom{{\downarrow}{1 }}$ &
${62.5} $$\phantom{{\downarrow}{1 }}$\\
 &\textbf{SALC+KGAT(ours)} & \phantom{0}${82.6} $$\phantom{{\downarrow}{1 }}$ & $\textbf{67.8} $$\phantom{{\downarrow}{1 }}$ & $\textbf{70.1} $$\phantom{{\downarrow}{1 }}$ & $\textbf{66.1} $$\phantom{{\downarrow}{1 }}$ &
$\textbf{64.5} $$\phantom{{\downarrow}{1 }}$ & 
$\textbf{65.3} $$\phantom{{\downarrow}{1 }}$ & $\textbf{50.0} $$\phantom{{\downarrow}{1 }}$ & $\textbf{58.2} $$\phantom{{\downarrow}{1 }}$ & ${65.4} $$\phantom{{\downarrow}{1 }}$ &
$\textbf{65.6} $$\phantom{{\downarrow}{1 }}$\\

\hline
\multirow{3}{*}{InfMAE}\phantom{0} & \text{STD} & \phantom{0}$\textbf{84.7}$$\phantom{{\downarrow}{1 }}$ & ${11.3} $$\phantom{{\downarrow}{1 }}$ & ${11.0} $$\phantom{{\downarrow}{1 }}$ & ${12.7} $$\phantom{{\downarrow}{1 }}$ & 
${12.4} $$\phantom{{\downarrow}{1 }}$ & ${17.5} $$\phantom{{\downarrow}{1 }}$ & ${58.8} $ $\phantom{{\downarrow}{1 }}$& ${49.8} $$\phantom{{\downarrow}{1 }}$ & ${44.8} $ $\phantom{{\downarrow}{1 }}$&
${33.7} $$\phantom{{\downarrow}{1 }}$\\
&\text{SALC} & \phantom{0}${76.3}$ $\phantom{{\downarrow}{1 }}$& ${68.3} $$\phantom{{\downarrow}{1 }}$ & ${66.9} $ $\phantom{{\downarrow}{1 }}$& ${67.1} $$\phantom{{\downarrow}{1 }}$ & 
${61.0} $$\phantom{{\downarrow}{1 }}$ & ${70.9} $$\phantom{{\downarrow}{1 }}$ & ${69.1} $$\phantom{{\downarrow}{1 }}$ & ${50.8} $$\phantom{{\downarrow}{1 }}$ & ${46.7} $$\phantom{{\downarrow}{1 }}$ &
${64.2} $$\phantom{{\downarrow}{1 }}$\\
&\textbf{SALC+KGAT(ours)} & \phantom{0}${80.4}$$\phantom{{\downarrow}{1 }}$ & $\textbf{72.4} $$\phantom{{\downarrow}{1 }}$ & $\textbf{70.9} $$\phantom{{\downarrow}{1 }}$ & $\textbf{68.9} $$\phantom{{\downarrow}{1 }}$ & 
$\textbf{64.6} $ $\phantom{{\downarrow}{1 }}$& $\textbf{75.1} $$\phantom{{\downarrow}{1 }}$ & $\textbf{64.3} $$\phantom{{\downarrow}{1 }}$ & $\textbf{54.4} $$\phantom{{\downarrow}{1 }}$ & $\textbf{50.9} $$\phantom{{\downarrow}{1 }}$ &
$\textbf{66.9} $$\phantom{{\downarrow}{1 }}$\\

\noalign{\vskip 0mm \hrule height 0.5mm \vskip 0mm}

\end{tabular}
}
\end{table*}

\textcolor{black}{ \noindent \textbf{Implementation Details.} In addition, our KGAT training method has good transferability and can be combined with other methods. Here we provide three different versions of KGAT: MTD-KGAT, SALC-KGAT, and RobustDet-KGAT, which are combined with one existing adversarial training method: MTD \citep{zhang2019mtd}, and other two robust architecture design methods: SALC \citep{Chen2024Accurate} and RobustDet \citep{Dong2022AdversariallyAware}.}

\textcolor{black}{Meanwhile, for models with NMS operations, we calculate the relative gray value relation based on the prediction boxes with the top 100 highest prediction probabilities for all types of classes before NMS. For models like DINO, which do not use NMS, we directly calculate the relative gray value relation based on the initial predictions of these models.}

\textcolor{black}{During training, all KGAT variants adopt a unified online PGD-based adversarial sampling scheme. We evaluate the trained models under PGD-based adversarial perturbations and additional unseen perturbations, including DAG attacks and common corruptions such as Gaussian noise, in order to assess the generalization of our method across diverse adversarial settings.}

\subsection{Robustness Performance} 
The performances of Faster R-CNN, YOLO-v8, DINO, and SSD trained by the baseline and KGAT, along with their performance under various attacks, are shown in Table \ref{main_experiment_M3FD} for M$^3$FD, Table \ref{main_experiment_flir_adav} for  FLIR-ADAS, and Table \ref{main_experiment_KAIST} for KAIST. 

\noindent \textbf{Adversarial Robustness.} From the results, KGAT has obviously improved adversarial robustness. Specifically, as shown in Table \ref{main_experiment_M3FD}, the SALC + KGAT method achieves adversarial robustness of 61.9\%, 69.9\%, and 69.8\%  mAP$_{50}$ for Faster R-CNN, YOLO-v8, and DINO against MTD attack on M$^3$FD, which are better than the SALC by 5.2\%, 8.6\%, and 6.1\%. Moreover, for the SSD, the RobustDet + KGAT method attains an mAP$_{50}$ of 38.6\% under the CWA attack, representing a 2.9\% improvement over the  RobustDet. 

\noindent \textbf{Robustness for Common Corruptions.} In addition, besides the adversarial attack, our KGAT method can obviously also improve the robustness against the common corruptions. Specifically, as shown in Table \ref{main_experiment_flir_adav}, on the FLIR-ADAS dataset, many methods experience varying degrees of accuracy degradation compared to the STD method when faced with salt noise, a common corruption, across the four infrared detectors. However, our SALC + KGAT and RobustDet + KGAT methods achieve accuracy improvements of 2.8\%, 4.3\%, 3.4\%, and 4.4\% in mAP$_{50}$ over the baseline methods (SALC and RobustDet), which already outperform the STD method on Faster R-CNN, YOLO-v8, DINO, and SSD, respectively, demonstrating the robustness of our KGAT against common corruptions.

\noindent \textbf{Clean Accuracy.} From the result, our KGAT achieves the most improvement in robustness with the least sacrifice in clean accuracy in different experimental settings. Even on the M$^3$FD and FLIR-ADAS datasets, our RobustDet+KGAT method achieves an improvement of 0.3\% and 0.5\% in accuracy on clean images for the SSD, respectively.

\noindent \textbf{Combination for Different Baseline Methods.} Furthermore, KGAT exhibits good compatibility and can be easily combined with different types of methods, regardless of training optimization (e.g., MTD) or architectural modification (e.g., SALC and RobustNet). Compared with the baseline methods, the different versions of KGAT: MTD+KGAT, SALC+KGAT, and RobustDet+KGAT, have achieved considerable robustness improvement under a wide variety of evaluation metrics. 

M$^3$FD and FLIR-ADAS contain a wide variety of classes, offering a wealth of relation knowledge for utilization. In contrast, the KAIST mainly consists of pedestrians and backgrounds. The performance on those different datasets validates the generalization of our KGAT in scenarios with a limited class number  and lower image resolution.

\noindent \textbf{Comparison with Re-weight based Methods.} Since our KGAT is the adversarial training with re-weight form, we select another re-weight adversarial method: CWAT for comparison. From the result, we can find that three variants of KGAT have better robust performance compared with CWAT. The results demonstrate that knowledge-guided re-weight ideology is a more ideal solution to enhance the robustness compared with data-driven re-weight ideology.

\noindent \textcolor{black}{\textbf{Performance Gap between Datasets.} KGAT imposes relative thermal-radiation relations as physical constraints in adversarial training, and its gain is closely related to the diversity of radiation patterns available in the training data. Since KAIST is essentially single-class (pedestrian), the learnable relations are mostly limited to pedestrian--background cues, which constrains the amount of stable and transferable radiation priors that KGAT can exploit. In contrast, multi-class datasets (e.g., M$^3$FD~\citep{liu2022target} with 6 categories) provide richer cross-class radiation patterns, yielding more evident robustness gains under adversarial attacks and common corruptions. For example, with the DINO detector on M$^3$FD, SALC+KGAT improves the \textbf{Avg} mAP$_{50}$ from 63.2 to 70.8 (+7.5), whereas on KAIST it increases from 63.2 to 69.3 (+6.1), resulting in a smaller gain. The results indicate that KGAT is more effective when there are more class relationships available.}

\textcolor{black}{
\noindent \textbf{Robustness on Specialized Infrared Detection architectures.} To verify the effectiveness of KGAT on specialized
infrared detection architectures, we select two typical methods: EFLNet \citep{yang2024eflnet} and InfMAE \citep{liu2024infmae}. Following the original setting in InfMAE, we adopt Mask R-CNN as the detection head when integrating InfMAE into the infrared object detection pipeline.
This setting strictly follows the recommended downstream configuration in the original work, ensuring a fair and faithful evaluation of KGAT on a strong infrared-specific backbone. All experiments are conducted on the M$^3$FD~\citep{liu2022target} dataset.}

\textcolor{black}{
 As shown in Table~\ref{detection_result}, KGAT consistently improves robustness for both infrared-specific models under adversarial attacks and common corruptions. \textbf{For EFLNet}, adding KGAT increases the average mAP$_{50}$ from 62.5 to 65.6; under adversarial attacks, it improves performance from 65.2 to 67.8 on $A_{cls}$ and from 66.8 to 70.1 on $A_{loc}$, while also yielding gains under common corruptions such as Blur (46.3 to 50.0). \textbf{For InfMAE}, despite its strong infrared-aware representations, KGAT still raises the average mAP$_{50}$ from 64.2 to 66.9; it improves robustness under $A_{cls}$ (68.3 to 72.4) and DAG (70.9 to 75.1), and also remains beneficial under noise corruptions (e.g., GN: 46.7 to 50.9). These results suggest that KGAT complements detection-level robustness training by injecting thermal-radiation priors, leading to more robust and transferable performance across diverse perturbations.
}

\textcolor{black}{
Overall, these results validate that KGAT is not restricted to a specific detector or backbone design, but can be seamlessly integrated into heterogeneous infrared object detection models, indicating its strong extensibility and practical applicability within the infrared domain.
}

\begin{table*}[t]
\renewcommand{\arraystretch}{1.3}
\setlength{\tabcolsep}{2.0pt}
\centering
\caption{Ablation Study towards the key components of our KGAT methods. The results are based on the M$^3$FD of YOLO-v8 and SSD. The best result is highlighted in bold, while the second-best result is marked with an underline.} 
\label{ablation study component}
\begin{tabular}{c|ccccc|ccccccccc}
\noalign{\vskip 0mm \hrule height 0.5mm \vskip 0mm}
       & MTD & SALC & RobustDet & $w_i^{\rho}$ & $w_i^{\mathcal{S}}$ & Clean & A$_{cls}$ & A$_{loc}$ & A$_{mtd}$ & CWA   & DAG  & Blur & Salt & GN   \\
\noalign{\vskip 0mm \hrule height 0.2mm \vskip 0mm}
\multirow{6}{*}{YOLO-v8} & $\checkmark$   &      &           &     &     & 65.8  & 57.2 & 56.3 & 55.9 & 54.5 & 56.1 & 41.0   & 51.7 & 46.8 \\
       & $\checkmark$   &      &           & $\checkmark$   &     & \underline{69.4}  & \underline{67.8} & \underline{63.8} & \underline{67.1} & \underline{64.2} & \underline{66.5} & \underline{46.4} & \underline{57.1} & \underline{53.3} \\
       & $\checkmark$   &      &           & $\checkmark$   & $\checkmark$   & \textbf{72.2}  & \textbf{68.4} & \textbf{67.7} & \textbf{68.8} & \textbf{66.5} & \textbf{67.1}   & \textbf{47.4} & \textbf{60.2} & \textbf{55.9} \\
\cmidrule{2-15}
       &     & $\checkmark$    &           &     &     & 71.2  & 62.1 & 60.7 & 61.3 & 59.8 & 60.1 & 43.9 & 55.1 & 51.2 \\
       &     & $\checkmark$    &           & $\checkmark$   &     & \underline{72.0}  & \underline{68.5} & \underline{64.2} & \underline{68.1} & \underline{66.9} & \underline{69.8} & \underline{51.6} & \underline{63.4} & \underline{58.9} \\
       &     & $\checkmark$    &           & $\checkmark$   & $\checkmark$   & \textbf{73.9}  & \textbf{70.6} & \textbf{67.1} & \textbf{69.9} & \textbf{67.5} & \textbf{71.2}   & \textbf{52.4} & \textbf{65.2} & \textbf{60.7} \\
\hline
\multirow{3}{*}{SSD}    &     &      & $\checkmark$         &     &     & 64.5  & 35.8 & 34.9 & 36.9 & 37.0   & 47.1 & 48.3 & 59.6 & 54.7 \\
       &     &      & $\checkmark$         & $\checkmark$   &     & \underline{71.4}  & \underline{39.9} & \underline{37.4} & \underline{37.2} & \underline{37.5}   & \underline{48.5} & \underline{48.5} & \underline{62.1} & \underline{57.1} \\
       &     &      & $\checkmark$         & $\checkmark$   & $\checkmark$   & \textbf{71.6}  & \textbf{42.0} & \textbf{39.7} & \textbf{37.6} & \textbf{38.6} & \textbf{49.1} & \textbf{48.6} & \textbf{63.8} & \textbf{58.8} \\
\noalign{\vskip 0mm \hrule height 0.5mm \vskip 0mm}
\end{tabular}
\end{table*}

\begin{figure}[t]
    \centering
    \includegraphics[width=1\linewidth]{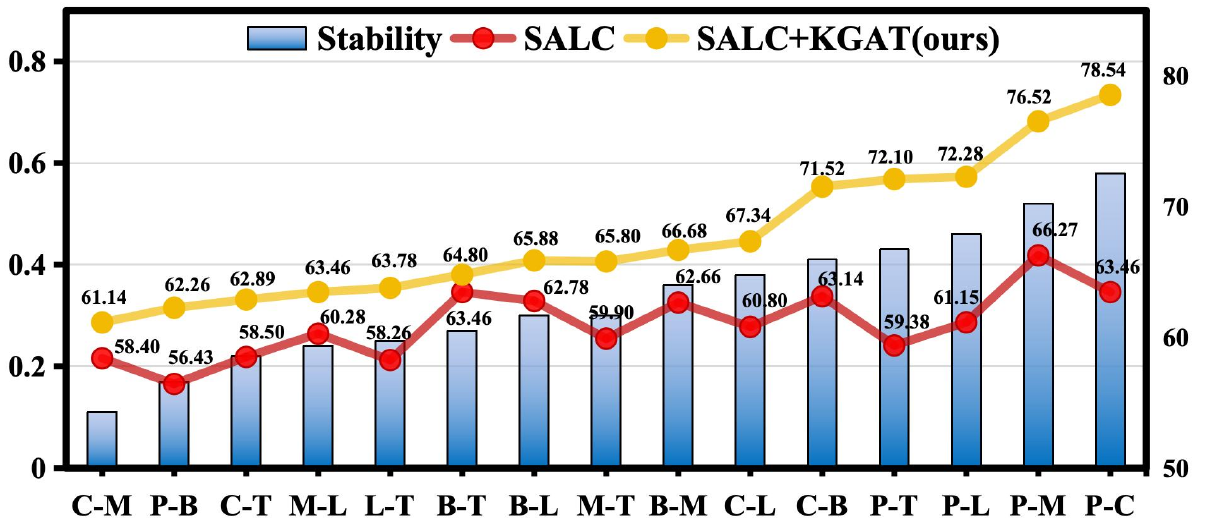}
    \caption{\textcolor{black}{The robustness of YOLO-v8 based SALC and SALC+KGAT (ours) against A$_{mtd}$ attack on images with thermal radiation relations of different stability. In the x-axis labels, C, P, B, M, L, and T denote Cars, People, Bus, Motorcycle, Lamp, and Truck, respectively. The x-axis denotes the specific thermal radiation relation, and the bars indicate the stability level of that relation, using the left y-axis. The two curves represent the mAP$_{50}$ of SALC and SALC+KGAT (ours) on images containing this specific thermal radiation relation, respectively, using the right y-axis.}}
    \label{fig:stability}
\end{figure}

\noindent \textcolor{black}{\textbf{Robustness for the Relation with Different Stability.} }\textcolor{black}{We explore the association between KGAT and the stability of infrared thermal radiation relation. Here we select all fifteen pairwise infrared thermal radiation relations, People-\{Cars, Bus, Motorcycle, Lamp, Truck\}, Cars-\{Bus, Motorcycle, Lamp, Truck\}, Bus-\{Motorcycle, Lamp, Truck\}, Motorcycle-\{Lamp, Truck\}, and Lamp-Truck, among the six categories on M$^{3}$FD using YOLO-v8 and report the robustness (mAP$_{50}$) against A$_{mtd}$ for images associated with each relation in Figure \ref{fig:stability}. We observe that our KGAT can not only obviously enhance the robustness for the images with stable thermal radiation relation, but can also slightly improve the robustness for images with unstable thermal radiation relation, which shows the superiority of our KGAT.}

\begin{figure}
    \centering
    \includegraphics[width=0.98\linewidth]{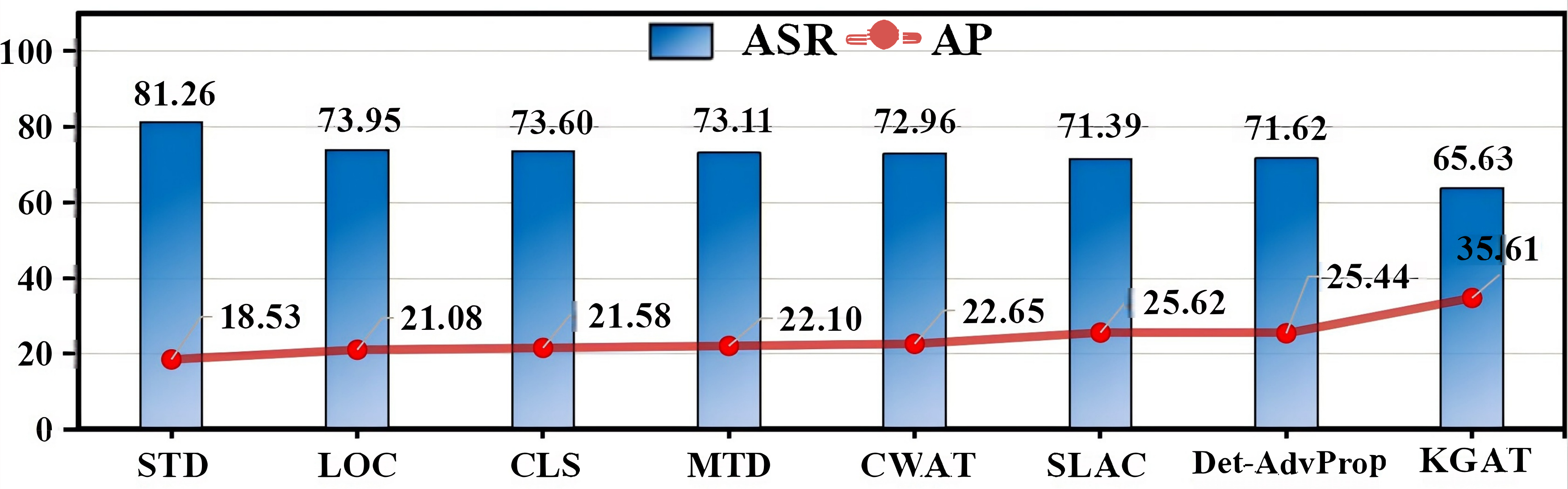}
    \caption{The robustness of YOLO-v8 against the AIP attack, with Average Precision (AP) plotted on the right Y-axis and Attack Success Rate (ASR) plotted on the left Y-axis. The KGAT is SALC+KGAT version.}
    \label{fig:patch attack}
\end{figure}

\noindent \textbf{Adversarial Robustness on Physical Attacks.}
Infrared object detection has shown that physical vulnerabilities \citep{zhu2021fooling, zhu2022infrared,wei2023physically,wei2023hotcold,hu2024adversarial}, and the robustness towards physical attacks should be carefully evaluated. Therefore, in this section, we select a representative physical attack: Adversarial Infrared Patches (AIP) attack \citep{wei2023physically} and investigate the robustness of our KGAT and other baseline methods.

Here we follow the dataset construction procedure in \cite{wei2023physically}, selecting the images that contain pedestrians from the FLIR-ADAS dataset as our dataset. As a result, the training set consists of 7,873 images, while the test set contains 2,027 images. Subsequently, we select the test images that can be successfully detected by the target model with high confidence as the final images on which attacks are performed. Therefore, the clean Average Precision (AP) is 100\%. We primarily adopt YOLO-v8 instead of YOLO-v3 in \cite{wei2023physically}, where YOLO-v8 offers higher accuracy and can fit more complex scenarios. Following \cite{wei2023physically} and \cite{hu2024adversarial}, we adopt Attack Success Rate (ASR) and Average Precision (AP) as the metrics to evaluate the effectiveness of adversarial attacks. 

The experimental results are shown in the Figure \ref{fig:patch attack}. Specifically, higher AP and lower ASR indicate stronger robustness of the object detector. Our SALC + KGAT method achieves the highest AP of 35.61\% and the lowest ASR of 65.63\%. Compared to the baseline method SALC, the AP increased by 9.99\%, and the ASR decreased by 5.76\%. In addition, the qualitative experimental results are shown in the Figure \ref{fig:image Amtd}. The above experimental results further demonstrate that the thermal radiation relation guided optimization and knowledge reliability aware optimization can obviously enhance the robustness of the object detector when faced with some underlying physical attacks such as the AIP attack. 

\noindent \textcolor{black}{\textbf{
Robust Detection across different IoU Thresholds.} }\textcolor{black}{To more comprehensively characterize the impact of our KGAT method, beyond the standard mAP$_{50}$, we additionally report mAP$_{75}$ and mAP$_{50:95}$ for YOLO-v8 on M$^{3}$FD under three settings: Clean, A$_{mtd}$, and Gaussian noise to provide a stricter assessment of bounding-box localization precision. The quantitative results are summarized in Table \ref{tab:detection_results}. The results show that our KGAT can outperform the baseline methods at different IoU thresholds. Specifically, our SALC + KGAT methods outperform SALC by 11.1\% and 9.9\% in the metric of mAP$_{75}$ and mAP$_{50:95}$ against $A_{mtd}$ Attack, showing the superiority of our KGAT. }
\begin{table}[h]
\centering
\caption{\textcolor{black}{Detection Performance across different IoU Thresholds, including mAP$_{75}$ and mAP$_{50:95}$. The results are based on the best checkpoints on $M^3FD$ of YOLO-v8.}}
\setlength{\tabcolsep}{2.0pt}
\label{tab:detection_results}
\begin{tabular}{l|ccc|ccc}
\toprule
\multirow{2}{*}{Method} &\multicolumn{3}{c|}{mAP$_{75}$} & \multicolumn{3}{c}{mAP$_{50:95}$} \\
\cmidrule(lr){2-4} \cmidrule(lr){5-7} 
 & Clean & $A_{mtd}$ & GN & Clean & $A_{mtd}$ & GN \\
\midrule
STD        &  \underline{56.3} & 7.6 & 34.4 & \textbf{54.1} & 7.4 & 33.0 \\
SALC       & 54.1 & \underline{41.9} & \underline{36.4} & 48.0 & \underline{39.0} & \underline{33.2} \\
\textbf{SALC+KGAT}   & \textbf{57.5} & \textbf{53.0} & \textbf{47.1} & \underline{53.2} & \textbf{48.9} & \textbf{43.7} \\
\bottomrule
\end{tabular}
\end{table}

In summary, extensive experiments have validated the effectiveness of our proposed KGAT method, further demonstrating that utilizing the infrared physical knowledge can obviously enhance the robustness of various object detectors in diverse complex scenarios.

\subsection{Ablation Study}
To certify the effectiveness of our method, we perform ablation experiments on every component. All the experiments are conducted based on M$^3$FD of YOLO-v8.

\begin{table}[t]
\renewcommand{\arraystretch}{1.3}
\setlength{\tabcolsep}{2.0pt}
\centering
\caption{Ablation study towards different mapping operations. The results of linear and nonlinear mapping are based on the best checkpoints on $M^3FD$ of YOLO-v8 based on SALC. The best result is highlighted in bold.} 
\label{ablation study1}
\begin{tabular}{c|cccccccc}
\noalign{\vskip 0mm \hrule height 0.5mm \vskip 0mm}
\multicolumn{1}{c}{Tpyes}      & Clean  & $A_{cls}$  & $A_{loc}$  & $A_{mtd}$   & DAG   & Blur  & Salt  & GN \\
\hline
Baseline    &  71.2 & 62.1 & 60.7 & 61.3 & 60.1 & 43.9 & 55.1 & 51.2    \\
Linear   & 71.4  & 69.4  & \textbf{68.5}   & 67.5  & 69.2  & 50.5  & \underline{64.5}  & 60.1     \\
Exp-based   & \underline{73.3}  & \underline{70.3}  &\underline{67.6}  & \underline{69.3}  & \underline{70.8}  & \underline{51.6}  & \underline{64.5}  & \underline{60.2}     \\
Log-based   & \textbf{73.9}  & \textbf{70.6}  & 67.1  & \textbf{69.9}  & \textbf{71.2}  & \textbf{52.4}  & \textbf{65.2}  & \textbf{60.7}     \\

\noalign{\vskip 0mm \hrule height 0.5mm \vskip 0mm}
\end{tabular}
\end{table}

\noindent \textbf{Effectiveness for different components of KGAT.}
KGAT consists of two key components: thermal radiation-guided optimization $w_i^{\rho}$, and infrared physical knowledge reliability aware optimization $w_i^{\mathcal{S}}$. The ablation experiments based on YOLO-v8 and SSD, conducted on the M$^3$FD dataset, are shown in  Table \ref{ablation study component}. 

From the results, we observe that the robustness of the models trained by integrating our KGAT method with various adversarial training methods is generally superior to that of the original adversarial training methods. Furthermore, the introduction of the infrared physical knowledge reliability aware $w_i^{\mathcal{S}}$ further enhances the model's performance. This is because $w_i^{\mathcal{S}}$ serves as a form of infrared physical knowledge that reflects the stability of various thermal radiation relations under complex interference scenarios, which restricts the model to focus more on learning thermal radiation relations that are robust to a range of adversarial attacks and common corruptions during the training phase, ultimately improving the overall accuracy and robustness of the model.

\noindent \textbf{Mathematical Form for $w_i^{\rho}$.} As for the ablation study, we initially perform ablation experiments towards the mathematical form for the $w_i^{\rho}$. In fact, as discussed in Section \ref{Thermal Radiation Relation Guided Optimization}, besides using the log-based method for nonlinear mapping, adopting a simpler nonlinear mapping form such as $w_i=e^{-\lambda\cdot\rho}$ can also achieve the desired effect, where $\lambda$ is a hyperparameter used to control the rate of change and $\rho$ is the correlation coefficient. Furthermore, a more direct option is to use a linear mapping form to convert the correlation coefficient into a weight. Therefore, here we explore the necessity of using a nonlinear mapping form and explain why we ultimately chose the log-based form for this nonlinear mapping. The results are shown in Table \ref{ablation study1}. All results are the optimal values under their respective mathematical forms of $w_i^{\rho}$.

Based on the results, we can find it is necessary to apply the mathematical form of nonlinear mapping for $w_i^{\rho}$. The nonlinear mapping applies more differentiated weights to samples with different order differences. Specifically, when $\rho$ decreases by a certain value $\delta$, the increase in $w_i^{\rho}$  is greater than the decrease in $w_i^{\rho}$ when $\rho$ increases by the same value $\delta$, allowing the model to focus more precisely on those samples with larger order differences, while linear changes do not have the corresponding effect. Additionally, we observe that exponential and logarithmic nonlinear mappings achieve similar results, but the logarithmic mapping consistently outperforms the exponential mapping. Therefore, we ultimately choose the logarithmic function as the mapping for $w_i^{\rho}$.

\noindent \textbf{Hyper-parameter Selection for $\beta$.} After discussing the mathematical form of $w_i^{\rho}$, we perform ablation experiments towards the value selection of hyper-parameter $\beta$, which can directly determine the knowledge-guided strength. And the results can be viewed in Table \ref{ablation study beta}.
From the results, we can find that the hyper-parameter $\beta$ can influence the final results. When the $\beta$ is large, the adjustment towards weights will become oscillatory, which may lead to unstable optimization results; when the $\beta$ is small, the adjustment towards weights will be weak, which may not play an obvious role in guidance. Finally, we select the final value of $\beta$ to be 0.5 in our experimental setting.

\begin{table}[t]
\renewcommand{\arraystretch}{1.3}
\setlength{\tabcolsep}{2.0pt}
\centering
\caption{Ablation Study towards the hyper-parameter $\beta$. The results are based on the M$^3$FD of YOLO-v8. The best result is highlighted in bold.} 
\label{ablation study beta}
\begin{tabular}{c|cccccccc}
\noalign{\vskip 0mm \hrule height 0.5mm \vskip 0mm}
$\beta $    & Clean & A$_{cls}$ & A$_{loc}$ & A$_{mtd}$ & DAG  & Blur & Salt & GN   \\
\hline
$\beta=0.3$ & 70.8  & 68.2 & 67.3 & 65.7 & 69.8 & 50.6 & 64.5 & 58.1 \\
$\beta=0.4$ & 71.7  & 69.3 & \textbf{67.6} & 66.5 & 70.4 & 50.7 & 64.6 & \underline{59.8} \\
$\beta=0.5$ & \textbf{73.9}  & \textbf{70.6} & 67.1 & \textbf{69.9} & \textbf{71.2}   & \textbf{52.4} & \textbf{65.2} & \textbf{60.7} \\
$\beta=0.6$ & \underline{72.8}  & 69.4 & \underline{67.5} & \underline{68.8} & 70.4 & \underline{51.9}   & \underline{65.0} & \underline{59.8} \\
$\beta=0.7$ & 71.7  & \underline{70.4} & 65.6 & 68.1 & \underline{70.9} & 51.6 & 64.7 & 58.1\\
\noalign{\vskip 0mm \hrule height 0.5mm \vskip 0mm}
\end{tabular}
\end{table}

\begin{table}[t]
\renewcommand{\arraystretch}{1.3}
\setlength{\tabcolsep}{2.0pt}
\centering
\caption{Ablation Study towards the hyper-parameter $\gamma$. The results are based on the M$^3$FD of YOLO-v8.  All attack settings are the same as in the previous experiments. The best result is highlighted in bold. } 
\label{ablation study gamma}
\begin{tabular}{c|cccccccc}
\noalign{\vskip 0mm \hrule height 0.5mm \vskip 0mm}
$\gamma$     & Clean & A$_{cls}$ & A$_{loc}$ & A$_{mtd}$ & DAG  & Blur & Salt & GN   \\
\hline
$\gamma=0.7$ & 69.5  & 67.4 & 64.8 & 67.4 & 69.7 & 47.8 & 59.5 & 58.1 \\
$\gamma=0.9$ & 72.6    & 69.3 & \textbf{68.5} & \underline{69.3} & 70.4 & 52.3 & 62.6 & \underline{60.4} \\
$\gamma=1.1$ & \textbf{73.9}  & \textbf{70.6} & 67.1 & \textbf{69.9} & \textbf{71.2}   & \underline{52.4} & \textbf{65.2} & \textbf{60.7} \\
$\gamma=1.3$ & \underline{72.8}  & \underline{70.3} & \underline{67.6} & 68.8 & \underline{70.6} & \textbf{54.0}   & 64.6 & 59.9 \\
$\gamma=1.5$ & 71.2  & 69.4 & 67.0 & 69.0 & 70.1 & 51.6 & \underline{65.0} & 59.0\\
\noalign{\vskip 0mm \hrule height 0.5mm \vskip 0mm}
\end{tabular}
\end{table}

\begin{table*}[t]
\centering
\caption{\textcolor{black}{ \textbf{Robustness comparison under imprecise bounding boxes.} We report results under three noise levels (10\%,20\%,30\%) and GN denotes Gaussian Noise. }}
\label{tab:box_noise}
\setlength{\tabcolsep}{6pt}
\begin{tabular}{l|ccc|ccc|ccc}
\hline
Noise Level& \multicolumn{3}{c|}{\textbf{10\%}} 
& \multicolumn{3}{c|}{\textbf{20\%}} 
& \multicolumn{3}{c}{\textbf{30\%}} \\ \hline
Method & Clean & A$_{mtd}$ & GN & Clean & A$_{mtd}$ & GN & Clean & A$_{mtd}$ & GN \\
\hline
STD & \textbf{75.1} & 9.1  & \underline{51.5} & \textbf{72.9} & 8.0  & \underline{44.3} & \textbf{69.8} & 6.1 & \underline{35.7} \\
SALC & 69.3 & \underline{56.3} & 50.8 & 67.1 & \underline{52.7} & 42.1 & 63.9 & \underline{49.3} & 34.8 \\
SALC+KGAT (ours) & \underline{72.1} & \textbf{64.3} & \textbf{58.3} & \underline{70.2} & \textbf{62.3} & \textbf{52.1} & \underline{67.8} & \textbf{59.3} & \textbf{44.3} \\
\hline
\end{tabular}
\end{table*}

\noindent \textbf{Hyper-parameter Selection for $\gamma$.}
Based on the discussion in Section \ref{Knowledge Reliability Aware Optimization}, $\gamma$ can be used to control the strength of knowledge reliability aware optimization. Here, we perform an ablation study on its value, and the results are shown in the Table \ref{ablation study gamma}. From the experimental results, we can observe that $\gamma$ obviously affects the final performance of the model. When $\gamma$ is small, its role in guidance is minimal; when $\gamma$ is large, the change in $w_i^{\mathcal{S}}$ becomes more pronounced, which could lead to an unstable training process in diverse and complex scenarios. Finally, we select the final value of $\gamma$ to be 1.1 in our experimental setting.

\subsection{Impact of Annotation Noise}

\textcolor{black}{Since our KGAT obtains the thermal radiation relation based on the annotation information, the annotation will inevitably have a negative impact towards the performance of our KGAT. To explore the negative impact produced by annotation noise, here we conduct a controlled box-perturbation study to empirically evaluate robustness under imprecise bounding boxes. Following the synthetic box-noise setting described in \cite{liu2022oamil}, we perturb each ground-truth box by random shifting and scaling. This design mimics realistic annotation inaccuracies, where the box center may drift and the box size may be inaccurately annotated, either larger or smaller than the true object region. We randomly perturb all ground-truth boxes in the training set for three different noise levels (10\%, 20\%, and 30\%) and train the object detector on the perturbed training set, while keeping all other training configurations unchanged. We report mAP$_{50}$ for YOLOv8 on M$^3$FD under three evaluation settings, including Clean, $A_{mtd}$, and Gaussian noise, and the results are summarized in Table~\ref{tab:box_noise}.
}

\textcolor{black}{ 
From the results, our KGAT achieves a notable improvement in robustness with minimal sacrifice in clean accuracy as the perturbation levels increase. Additionally, as the perturbation magnitude grows, the performance gap between SALC and SALC+KGAT becomes more evident when faced with $A_{mtd}$ and Gaussian noise. Specifically, when the perturbation level $r$ is set to 10\%, SALC+KGAT outperforms SALC by 7.5\% under Gaussian noise (58.3\% vs. 50.8\%). As the perturbation level increases to 30\%, the performance gain reaches 9.5\% (44.3\% vs. 34.8\%). These results suggest that our method exhibits enhanced robustness to imprecise bounding boxes, demonstrating its effectiveness in addressing annotation inaccuracies.
}

\subsection{Computational Overhead}

\textcolor{black}{Here we report the computational overhead introduced by KGAT during adversarial training. Specifically, we include: (i) the average per-epoch training time, computed by averaging the total training time over 100 epochs; (ii) the computational complexity measured by GFLOPs; and (iii) the peak GPU memory usage measured with a batch size of 8.  These computational overhead measurements are based on the YOLO-v8 and the M$^3$FD dataset. All other training settings follow the experimental setup in Section 5.1 for a fair comparison.}

\begin{table}[t]
\renewcommand{\arraystretch}{1.2}
\setlength{\tabcolsep}{2pt}
\centering
\caption{Computational Overhead. Average per-epoch training time with overall GFLOPs and GPU memory.}
\label{tab:overhead}
\begin{tabular}{lccc}
\hline
Methods & Time(s) & GFLOPs & Memory(GB) \\
\hline
MTD&306.98  &126.88  &22.30 \\
CWAT &362.36 & 131.26& 22.68 \\
Det-AdvProp &342.26 & 132.27& 23.58  \\
SALC & 313.31 & 129.07  & 22.69\\
\hline
\textbf{MTD+KGAT(ours)} &316.23 &128.05 &22.36 \\
\textbf{SALC+KGAT(ours)} & 325.46 & 130.24 &22.74  \\
\hline
\end{tabular}
\end{table}

\textcolor{black}{
The additional computational overhead of our SALC+KGAT and MTD+KGAT methods, compared to SALC and MTD, primarily stems from the integration of thermal radiation knowledge constraints into the model's training process. Specifically, as shown in Table~\ref{tab:overhead},  compared to SALC, SALC+KGAT results in a marginal increase in per-epoch training time (from 313.31s to 325.46s, an increase of approximately 3.88\%), and a slight rise in computational cost (from 129.07 to 130.24 GFLOPs, about 0.91\%). The GPU memory consumption remains nearly unchanged (22.69GB vs. 22.74GB, a 0.22\% increase). Similarly, compared to MTD, the increase in computational overhead for MTD+KGAT is of a similar magnitude.
}

\textcolor{black}{
However, it is important to emphasize that the computational overhead of our method remains relatively modest compared to other methods, such as CWAT and Det-AdvProp, which incur significantly higher computational costs. This confirms that, despite the added complexity of incorporating infrared physical knowledge, the performance improvements are achieved with minimal increases in computational cost.}

\textcolor{black}{
As for \textbf{inference speed}, KGAT only optimizes the adversarial training procedure and does not change the model architecture or inference pipeline. Therefore, it introduces no additional inference-time overhead. 
}


\section{Limitation} \label{sec:limitation}
\textcolor{black}{While our KGAT effectively improves the robustness of object detection models, the improvement is not obvious for scenarios where thermal radiation relations are not particularly stable. Furthermore, in extreme scenarios outside our statistical scope, such as large-area occlusion by hot objects or changes in object grayscale relationships due to extreme ambient temperatures or excessively strong adversarial and common corruptions, the performance of our KGAT is inevitably affected to some extent. Therefore, incorporating ambient temperature and other cues as conditions into the object detection model is a worthwhile direction to explore in future research.  Furthermore, when the number of available training data categories is limited, the available thermal radiation relationships for the method will decrease, and the magnitude of the method's improvement will also decline. Meanwhile, although our KGAT achieves the most improvement in robustness with the least sacrifice in clean accuracy, the trade-off between accuracy and robustness still exists, requiring further exploration in the future.}

\section{Conclusion} \label{sec:Conclusion}

This paper revisited the inherent infrared physical knowledge in infrared images and discovered that relative thermal radiation relations could serve as a stable source of knowledge. Consequently, we modeled the thermal radiation relations based on the rank order of different classes and quantified the stability of this knowledge based on the variations in these relations. Based on this theoretical framework, we embedded thermal radiation relations into the adversarial training framework for infrared object detection, which guided the model's optimization process and rendered the model's predictions more consistent with infrared physical principles. Extensive experiments demonstrated that our Knowledge-Guided Adversarial Training (KGAT) method effectively improved the robustness of infrared object detection against adversarial attacks and common corruptions across various datasets and detection models. KGAT explored a stable path for leveraging physical properties in infrared visual recognition and demonstrated broad application prospects for the future.


\section*{Data Availability Statements}
Information on access to the datasets supporting the conclusions of this article is included therein.

\section*{Conflict of Interest}
The authors declare that they have no conflict of interest.

\clearpage
\begin{appendices}

\section{Theoretical proof}
\subsection{The Proof of Corollary 1}
\textcolor{black}{
For the $k$-th class, the object-side spectral radiance is characterized by a gray-body model, where the spectral radiance $L_k^{\mathrm{obj}}(\lambda,u)$ equals the product of the emissivity and the Planck blackbody spectral radiance \cite{Modest2013RadiativeHeatTransfer,Planck1901NormalSpectrum}:
}

\textcolor{black}{
\begin{equation}
\label{eq:plank}
L_k^{\mathrm{obj}}(\lambda,u)=\varepsilon_k(\lambda,u)\,B(\lambda,T_k(u)),
\end{equation}
where $\lambda$ denotes the wavelength, and $u$ indexes a spatial point on the object surface.
$T_k(u)$ represents the absolute temperature of a class-$k$ object at location $u$, allowing for spatially varying temperature over the surface.
$\varepsilon_k(\lambda,u)\in(0,1]$ is the spectral emissivity at $(\lambda,u)$, which may depend on both wavelength and surface position and quantifies the departure of the object from an ideal blackbody.
Accordingly, $L_k^{\mathrm{obj}}(\lambda,u)$ is the emitted object-side spectral radiance at wavelength $\lambda$ from location $u$, and $B(\lambda,T_k(u))$ is the Planck blackbody spectral radiance evaluated at temperature $T_k(u)$. We treat $\varepsilon_k$ and $T_k$ as class-conditional variables to capture class differences at a statistical level.
After propagation, the at-sensor spectral radiance $L_k^{\mathrm{sen}}(\lambda,u)$ is:
}

\textcolor{black}{
\begin{equation}
\label{eq:sensor}
L_k^{\mathrm{sen}}(\lambda,u)=\tau(\lambda,u)\,L_k^{\mathrm{obj}}(\lambda,u)+L^{\mathrm{path}}(\lambda,u),
\end{equation}
where $\tau(\lambda,u)\ge 0$ and path radiance $L^{\mathrm{path}}(\lambda,u)$ jointly determine the received radiance.
}

\textcolor{black}{
Let $S(\lambda)\ge 0$ be the system spectral response. The band-integrated effective radiance $\bar L_k(u)$ is:
}

\textcolor{black}{
\begin{equation}
\label{eq:Lk}
\bar L_k(u)=\int_{\Lambda}S(\lambda)\,L_k^{\mathrm{sen}}(\lambda,u)\,d\lambda.
\end{equation}
}

\textcolor{black}{
We unify the infrared imaging and post-processing pipeline as a single operator $F$. Following the  imaging response modeling in \cite{Shi2005FeasibleNUC}, we assume $F$ is approximately monotonically increasing with respect to the effective radiance input, i.e., stronger effective radiance typically leads to a stronger gray value response in the final output image:
}

\textcolor{black}{
\begin{align}
\label{eq:xk}
y_k(u)=F(\bar L_k(u))+\delta(u),
\end{align}
where $y_k(u)$ denotes the gray value of the $k$-th class object at spatial location $u$ in the final output image. Since $F$ is approximately monotonically increasing with respect to its input, it maps locations with stronger thermal radiation signals to higher gray values, which are then used to construct classwise statistics and the subsequent training constraints. The perturbation terms $\delta(u)$  denotes the class-agnostic gray
value estimation error at location $u$, which arises from
environment noise or adversarial perturbation.
}



\textcolor{black}{
Based on the analysis above, in the $i$-th image, we extract the $k$-th by the mean gray value $\mathcal{G}_k^{(i)}$ over its annotated region as follows: 
\begin{align}
\label{eq:average_gray_value}
\mathcal{G}_k^{(i)}
&=\frac{1}{|\Omega_k^{(n)}|}\sum_{u\in\Omega_k^{(n)}} y_{k}^{(n)}(u) \notag \\
&=  \mathbb{E}_{u\sim\Omega_k^{(n)}}\!\big[F(\bar L_k(u))\big]+\mathbb{E}_{u\sim\Omega_k^{(i)}}\!\big[\delta(u)\big],
\end{align}
where $\mathbb{E}_{u\sim\Omega_k^{(n)}}[\cdot]$ denotes the uniform spatial average over pixel locations $u$ within the region $\Omega_k^{(n)}$.
}''

\subsection{The Proof of Theorem 1}
\label{proof1}

In an infrared scene dataset $\mathcal{C}$ that contains
$N$ images, $(x_i,y_i)\in \mathcal{C}$ is a sample in the dataset, where $x_i$ represents an infrared image and $y_i$ represents the corresponding annotations. The initial parameter of the infrared object detector $M_{\theta}$ is $\theta$.  $\theta_1$ denotes the new parameter updated with the loss function $\mathcal{L}_{det}$, while $\theta_2$ denotes the new parameter updated with the  $\mathcal{L}'_{KGAT}$. Then, $\theta_1$ and $\theta_2$ can be written as:

\begin{align}
\label{theta_1_2}
\theta_1 &= \theta-\eta\cdot\frac{1}{N}\sum_{i=1}^{N}\nabla \mathcal{L}_{det}(M_{\theta}(x_i),y_i), \\
\theta_2 &= \theta-\eta\cdot\frac{1}{N}\sum_{i=1}^{N}\nabla \mathcal{L}'_{KGAT}(M_{\theta}(x_i),y_i).
\end{align}

For $\forall (x,y)\in\mathcal{C}$, assume that the set of all foreground bounding boxes predicted by the infrared object detector $M_\theta$ on $x$ is denoted as $\mathcal{B}$. For $\forall b\in\mathcal{B}$, let $\mathcal{P}_i(b)$ denote the probability that the foreground bounding box $b$ is predicted to belong to the $i$-th class. Let $\mathcal{F}_0$ and $\mathcal{F}_i$ denote the gray value distributions of the entire image $x$ and $i$-th class, respectively, and their means are denoted as $\mu_0^x$ and $\mu_i^x$, respectively. Let $\{\mathcal{B}_1,\mathcal{B}_2,\ldots,\mathcal{B}_K\}$ be a partition of $\mathcal{B}$, for a specific $\mathcal{B}_i$, all foreground bounding boxes contained within it are assigned the ground truth class $i$. Then for $\forall b_j^{(i)}\in\mathcal{B}_i$, its gray value follow the distribution:

\begin{align}
\label{distrubition}
b_j^{(i)} \sim & (1-P_i(b_j^{(i)}))\cdot\mathcal{F}_0+\notag\\  &P_i(b_j^{(i)})\cdot[\frac{\Tilde{S}_j^{(i)}}{S_j^{(i)}}\cdot\mathcal{F}_i+\frac{S_j^{(i)}-\Tilde{S}_j^{(i)}}{S_j^{(i)}}\cdot\mathcal{F}_0],
\end{align}
where $S_j^{(i)}$ denotes the total area of the foreground bounding box 
$b_j^{(i)}$, and $\Tilde{S}_j^{(i)}$ denotes the overlapping area between $b_j^{(i)}$ and its corresponding ground truth.

Therefore, based on the predicted bounding boxes, the average gray value of the $i$-th class in the image $x$ can be calculated as:

\begin{align}
\label{gray value}
\mathcal{G}_i^x = \mu_0^x+(\mu_i^x-\mu_0^x)\cdot E[P_i]\cdot E[\frac{\Tilde{S}^{(i)}}{{S}^{(i)}}],
\end{align}
where:
\begin{align}
\label{Expectation_P}
E[P_i] = \frac{1}{\mid\mathcal{B}_i\mid}\sum_{j=1}^{\mid\mathcal{B}_i\mid}\mathcal{P}_i(b_j^{(i)}),
\end{align}
\begin{align}
\label{Expectation_S}
E[\frac{\Tilde{S}^{(i)}}{{S}^{(i)}}] =\frac{1}{\mid\mathcal{B}_i\mid} \sum_{j=1}^{\mid\mathcal{B}_i\mid}\frac{\Tilde{S}_j^{(i)}}{S_j^{(i)}}.
\end{align}

For simplicity, we denote: $E_i=E[P_i]\cdot E[\frac{\Tilde{S}^{(i)}}{{S}^{(i)}}]$. $E_i$ reflects the overall performance of model $M_\theta$, a larger $E_i$ indicates that model $M_{\theta}$ has better overall detection performance for the $i$-th class. Here, we assume that $E_i$ remains unchanged across different images.

Therefore, for any two classes, the $i$-th class and the $j$-th class in image $x$, the difference between their gray values is:

\begin{align}
\label{GV_diff}
\mathcal{G}_i^x-\mathcal{G}_j^x=E_i\cdot(\mu_i^x-\mu_0^x)-E_j\cdot(\mu_j^x-\mu_0^x).
\end{align}

For different images $x$, the gray value means $\mu_0^x$ and $\mu_i^x$ of image $x$ and class $i$ vary across images. However, based on previous analyses, we know that the relative magnitude relation between the gray values remains stable. Therefore, we assume the following relationship holds:

\begin{align}
\label{miu_dis}
\mu_i^x-\mu_0^x \sim \mathcal{N}(\Tilde{\mu}_i,\sigma_i^2),
\end{align}
where a larger absolute value of $\Tilde{\mu}_i$ and a smaller $\sigma_i^2$ indicate a more stable relationship.

Therefore, $X(i,j) = \mathcal{G}_i^x-\mathcal{G}_j^x$ is a random variable, it follows the distribution:

\begin{align}
\label{Xij}
X(i,j) \sim \mathcal{N}(E_i\Tilde{\mu}_i-E_j\Tilde{\mu}_j,  E_i^2\sigma_i^2+E_j^2\sigma_j^2).
\end{align}

For simplicity, we denote $\mathcal{P}_{\theta}(X(i,j)>0)$ as $\mathcal{P}_{\theta}(X(i,j))$, which represents the probability that the gray value of the $i$-th class is greater than that of the $j$-th class among the prediction results of model $M_{\theta}$. Thus,  $\mathcal{P}_{\theta}(X(i,j))$ can be calculated as follows:

\begin{align}
\label{P_Xij}
\mathcal{P}(X(i,j)) &= 1 - \Phi\left( \frac{-(E_i\tilde{\mu}_i - E_j\tilde{\mu}_j)}{\sqrt{E_i^2\sigma_i^2 + E_j^2\sigma_j^2}} \right)\notag \\ 
&= \Phi\left( \frac{\frac{E_i}{E_j} \cdot \tilde{\mu}_i - \tilde{\mu}_j}{\sqrt{\left( \frac{E_i}{E_j} \right)^2 \cdot \sigma_i^2 + \sigma_j^2}} \right),
\end{align}

where $\Phi(x)$ is the cumulative distribution function (CDF) of the standard normal distribution.

Next, we denote $\{\mathcal{C}_1,\mathcal{C}_2,\ldots,\mathcal{C}_l\}$ is a partition of $\mathcal{C}$, which means $\cup_{i=1}^{l}\mathcal{C}_i=\mathcal{C}$,  and for $\forall 1\leq i,j\leq l$ with $i\neq j$, it holds that $\mathcal{C}_i \cap \mathcal{C}_j=\emptyset$. Meanwhile, for a specific $\mathcal{C}_i$, all images $x$ contained in it have exactly $k_i$ distinct classes. Let $S_{k_i}$ be a symmetric group which is defined as the set of all bijections from the set $\{1,2,\ldots,k_i\}$ to itself, that is:
\begin{equation}
\resizebox{\columnwidth}{!}{$
\begin{aligned}
\label{S_ki}
S_{k_i}=\{\tau\mid\tau:(1,2,\ldots,k_i)\to (\tau(1),\tau(2),\ldots,\tau(k_i))\},
\end{aligned}
$}
\end{equation}
where $\tau$ is a permutation that rearranges the elements of the set $\{1,2,\ldots,k_i\}$. For simplicity, let us denote:
\begin{align}
\label{X_tau}
X_{\tau}(i)=X(\tau(i+1),\tau(i)),
\end{align}
where $\tau$ is a permutation, and $X(i,j)$ is defined in Eq.(\ref{Xij}).  $X_{\tau}(i)$ represents the probability that the gray value of the $\tau(i+1)$-th class is greater than that of the $\tau(i)$-th class in dataset $\mathcal{C}_i$.

Therefore, according to the chain rule, the expectation $E^i_{\theta}[\rho]$ of Spearman  rank correlation coefficient $\rho^{i}$ for the prediction results of model $M_\theta$ on dataset $\mathcal{C}_i$ can be calculated as follows:

\begin{align}
\label{E_S}
E^i_{\theta}[\rho]&=\frac{1}{Z(S_{k_i})}\sum_{\tau\in S_{k_i}}\prod_{i=1}^{k_i-1}\mathcal{P}_{\theta}(\tau(i))\cdot\rho^{i}_{\tau},\\
Z(S_{k_i})&=\sum_{\tau\in S_{k_i}}\prod_{i=1}^{k_i-1}\mathcal{P}_{\theta}(\tau(i)),\\
\mathcal{P}_{\theta}(\tau(i)) &= \mathcal{P}_{\theta}\left(X_{\tau}(i) \middle| X_{\tau}(1), \ldots, X_{\tau}(i-1) \right),
\end{align} 
where $\rho^{i}_{\tau}$ denotes the Spearman rank correlation coefficient between $(\tau(1),\tau(2),\ldots,\tau(k_i))$ and \allowbreak $(1,2,\ldots,k_i)$,  $Z(S_{k_i})$ is the partition function which is to normalize the probability distribution, and $\mathcal{P}_{\theta}(\tau(i))$ is a conditional probability.

Compared with $\theta_1$, $\theta_2$ is the result obtained after model $M_{\theta}$  undergoes gradient update using $\mathcal{L}'_{KGAT}$ as the loss function. This enables the prediction results of model $M_{\theta_2}$ to be more consistent with the actual thermal radiation characteristics of the objects. In other words, compared with $M_{\theta_1}$, $M_{\theta_2}$ achieves better detection performance $E_i$ for the $i$-th class, which has a greater gray value difference from the background. Under this condition, we assume the following relationship holds:

\begin{align}
\label{E_com}
(\frac{E_i}{E_j})_{\theta_2}>(\frac{E_i}{E_j})_{\theta_1}, \text{ if } \Tilde{\mu}_i > \Tilde{\mu}_j,
\end{align}
where $(\frac{E_i}{E_j})_{\theta_2}$ and $(\frac{E_i}{E_j})_{\theta_1}$ denote the ratio of the detection performance of the $i$-th class to that of the $j$-th class on model $M_{\theta_2}$ and model $M_{\theta_1}$, respectively.

Next, we proceed to prove that:
\begin{align}
\label{E_diff}
E^i_{\theta_2}[\rho]>E^i_{\theta_1}[\rho].
\end{align}

\textbf{First}, we take the pair $(i,j)$ that satisfies $1\leq j<i\leq k_i$. Based on it, we partition the symmetric group $S_{k_i}$ as follows:

\begin{align}
\label{Ski_partition}
\notag S_{k_i}^1 =& \{\tau\in S_{k_i}\mid \exists m\in\{1,2,\ldots,k_i-1\}, \\ &\tau(m)=j \text{ and } \tau(m+1)=i\},\\
\notag S_{k_i}^2 =& \{\tau\in S_{k_i}\mid  \exists m\in\{1,2,\ldots,k_i-1\},\\ &
\tau(m)=i \text{ and } \tau(m+1)=j\},\\
\Tilde{S}_{k_i} =& S_{k_i} \setminus (S_{k_i}^1\cup S_{k_i}^2).
\end{align}

By the definitions of $S_{k_i}^1$ and $S_{k_i}^2$, for $\forall \tau_1\in S_{k_i}^1$, $\exists \tau_2\in S_{k_i}^2$ and $m\in\{1,2,\ldots,k_i-1\}$, such that $\tau_1(l)=\tau_2(l)$ holds for all $l\neq m,m+1$, and $\tau_1(m)=\tau_2(m+1)=j$, $\tau_1(m+1)=\tau_2(m)=i$. Correspondingly, for $\forall \tau_2\in S_{k_i}^2$, $\exists \tau_1\in S_{k_i}^1$ that satisfies the aforementioned relationship. Then Eq.(\ref{E_S}) with parameter $\theta_1$ can be rewritten as: 

\begin{align}
\label{Ei1}
\notag E^i_{\theta_1}[\rho] =&\frac{1}{Z(S_{k_i})}\sum_{\tau\in S_{k_i}^1}\prod_{i=1}^{k_i-1}\mathcal{P}_{\theta_1}(\tau(i))\cdot\rho^{i}_{\tau}+\\ \notag &\frac{1}{Z(S_{k_i})}\sum_{\tau\in S_{k_i}^2}\prod_{i=1}^{k_i-1}\mathcal{P}_{\theta_1}(\tau(i))\cdot\rho^{i}_{\tau}+\\ 
&\frac{1}{Z(S_{k_i})}\sum_{\tau\in \Tilde{S}_{k_i}}\prod_{i=1}^{k_i-1}\mathcal{P}_{\theta_1}(\tau(i))\cdot\rho^{i}_{\tau}.
\end{align}

Based on Eq.(\ref{P_Xij}), let $E^i_{\theta_1}[\rho]_1$ denote the result obtained by replacing all instances of $(\frac{E_i}{E_j})_{\theta_1}$ and $(\frac{E_j}{E_i})_{\theta_1}$ in $E^i_{\theta_1}[\rho]$ with $(\frac{E_i}{E_j})_{\theta_2}$ and $(\frac{E_j}{E_i})_{\theta_2}$, respectively. All other parts keep unchanged, then based on Eq.(\ref{P_Xij}) and Eq.(\ref{E_com}), we have:

\begin{equation}
\resizebox{\columnwidth}{!}{$
\begin{aligned}
\label{com_PXij}
&\mathcal{P}_{\theta_2}(X(i,j))-\mathcal{P}_{\theta_1}(X(i,j))  \\  &=\Phi\left( \frac{(\frac{E_i}{E_j})_{\theta_2} \cdot \tilde{\mu}_i - \tilde{\mu}_j}{\sqrt{\left( \frac{E_i}{E_j} \right)_{\theta_2}^2 \cdot \sigma_i^2 + \sigma_j^2}} \right)-\Phi\left( \frac{(\frac{E_i}{E_j})_{\theta_1} \cdot \tilde{\mu}_i - \tilde{\mu}_j}{\sqrt{\left( \frac{E_i}{E_j} \right)_{\theta_1}^2 \cdot \sigma_i^2 + \sigma_j^2}} \right) \\  &=-\Phi\left( \frac{(\frac{E_j}{E_i})_{\theta_2} \cdot \tilde{\mu}_j - \tilde{\mu}_i}{\sqrt{\left( \frac{E_j}{E_i} \right)_{\theta_2}^2 \cdot \sigma_j^2 + \sigma_i^2}} \right)+\Phi\left( \frac{(\frac{E_j}{E_i})_{\theta_1} \cdot \tilde{\mu}_j - \tilde{\mu}_i}{\sqrt{\left( \frac{E_j}{E_i} \right)_{\theta_1}^2 \cdot \sigma_j^2 + \sigma_i^2}} \right) \\  
&= -\left(\mathcal{P}_{\theta_2}(X(j,i))-\mathcal{P}_{\theta_1}(X(j,i))\right)>0.
\end{aligned}
$}
\end{equation}

Take $\tau_1\in S_{k_i}^{1}$ and $\tau_2\in S_{k_i}^{2}$ that satisfy the aforementioned conditions. By the definition of the Spearman rank correlation coefficient, the following relationship holds:

\begin{align}
\label{com_rho}
\rho^i_{\tau_1}<\rho^i_{\tau_2}.
\end{align}

Thus, based on Eq.(\ref{com_PXij}), Eq.(\ref{com_rho}) and Eq.(\ref{Ei1}),  the following inequality holds:

\begin{align}
\label{com_Ep1Ep2}
E^i_{\theta_1}[\rho]_1>E^i_{\theta_1}[\rho].
\end{align}

\textbf{Then}, we iterate over all pairs $(i,j)$, where $1\leq j<i\leq k_i$ and $i\neq j$, in sequence following the calculation steps described above. In other words, in each iteration, we replace $(\frac{E_i}{E_j})_{\theta_1}$ and $(\frac{E_j}{E_i})_{\theta_1}$ with $(\frac{E_i}{E_j})_{\theta_2}$ and $(\frac{E_j}{E_i})_{\theta_2}$, respectively, where the pair $(i,j)$ has not yet been taken in the previous calculation result, until all $(i,j)$ pairs are taken. Let the total number of iterations be denoted as $K$, thus we can obtain the following inequality:

\begin{align}
\label{E_neq_scequence}
E^i_{\theta_1}[\rho]_K>E^i_{\theta_1}[\rho]_{K-1}>\ldots>E^i_{\theta_1}[\rho]_1>E^i_{\theta_1}[\rho].
\end{align}

Since all instances of $(\frac{E_i}{E_j})_{\theta_1}$ in $E^i_{\theta_1}[\rho]_K$ have already been replaced with $(\frac{E_i}{E_j})_{\theta_2}$, we have $E^i_{\theta_1}[\rho]_K=E^i_{\theta_2}[\rho]$. Thus, based on Eq.(\ref{E_neq_scequence}), Eq.(\ref{E_diff}) holds. Since Eq.(\ref{E_diff}) holds for any partition $\mathcal{C}_{i}$ of $\mathcal{C}$, on the entire dataset $\mathcal{C}$, we have:

\begin{align}
\label{Ep2>Ep1}
E_{\theta_2}[\rho]>E_{\theta_1}[\rho].
\end{align}

Finally, due to:
\begin{align}
\label{last}
\notag E_{\theta_h}[\mathcal{L}_{knowledge}] &=E_{\theta_h}[1-\rho]\\ &=1-E_{\theta_h}[\rho],h\in\{1,2\},
\end{align}
we have $E_{\theta_1}[\mathcal{L}_{knowledge}]>E_{\theta_2}[\mathcal{L}_{knowledge}]$. 

\textbf{This completes the proof.}




\section{Stability Analysis}


\textcolor{black}{
Under a fixed scene $S$, we model the class-wise gray-value statistic (e.g., the mean gray value inside the object box) as random variables $\mathcal{G}_{k}$ and $\mathcal{G}_{\Tilde{k}}$. Following the Gaussian statistical assumption in \cite{ge2020klgaussian_ir} and the Gaussian modeling of region appearance in \cite{rother2004grabcut}, and for tractable analysis, we simplify and assume that they follow single Gaussian distributions:
}

\textcolor{black}{
\begin{equation}
\label{eq:gaussian_distribution_XA_XB}
\mathcal{G}_{k} \sim \mathcal N(\mu_{k},\sigma_{k}^2),\qquad
\mathcal{G}_{\Tilde{k}} \sim \mathcal N(\mu_{\Tilde{k}},\sigma_{\Tilde{k}}^2).
\end{equation}
}

\textcolor{black}{
To characterize the ordinal relation between these two classes under scene $S$, we introduce the difference random variable $Z$:
}

\textcolor{black}{
\begin{equation}
\label{eq:Z_(valuable)}
Z = \mathcal{G}_{k} - \mathcal{G}_{\Tilde{k}},
\end{equation}
where $Z>0$ indicates that the $k$-th class  has a larger gray value statistic than the $\Tilde{k}$-th class in the $i$-th image, whereas $Z<0$ indicates the opposite.
}

\textcolor{black}{
We further assume that, within the same image, the samples from the two classes are independent, i.e., $\mathcal{G}_{k}$ and $\mathcal{G}_{\Tilde{k}}$ are independent. In fact, during infrared imaging, the mutual interference between the thermal radiation of two classes on each other’s imaging can be considered negligible. Then $Z$ is still Gaussian, with mean and variance given by:
}
\textcolor{black}{
\begin{align}
\label{eq:Z_gaussian}
Z \sim \mathcal N(\mu_\Delta,\sigma_\Delta^2),
\mu_\Delta=\mu_{k}-\mu_{\Tilde{k}},
\sigma_\Delta^2=\sigma_{k}^2+\sigma_{\Tilde{k}}^2.
\end{align}
}

\textcolor{black}{
We next derive the probability that $Z$ is positive (i.e. $\mathbb P(Z>0)$). According to Eq.\ref{eq:Z_gaussian}, the right-tail probability beyond zero is:
}

\textcolor{black}{
\begin{equation}
\label{eq:P(Z>0)}
\mathbb P(Z>0)=1-\Phi\!\left(-\frac{\mu_\Delta}{\sigma_\Delta}\right),
\end{equation}
where $\Phi(\cdot)$ denotes the CDF of the standard normal distribution.  Furthermore, we define a sign variable $Y$ as follows:
}

\textcolor{black}{
\begin{equation}
\label{Y_i_and_Z_i}
Y=\operatorname{sgn}(Z).
\end{equation}
}

\textcolor{black}{
Since Gaussians are continuous, $\mathbb P(Z=0)=0$, and thus $Y$ can be treated as a binary variable taking values in $\{-1,1\}$. Moreover, $Y=1$ corresponds to  $Z>0$ and $Y=-1$ corresponds to  $Z<0$. Therefore, the probabilities of the two outcomes are:
}

\textcolor{black}{
\begin{equation}
\label{P_Y_{i}>0}
\mathbb P(Y=1)=\mathbb P(Z>0)=1-\Phi\!\left(-\frac{\mu_\Delta}{\sigma_\Delta}\right),
\end{equation}
}

\textcolor{black}{
\begin{equation}
\label{P_Y_{i}<0}
\mathbb P(Y=-1)=\mathbb P(Z<0)=\Phi\!\left(-\frac{\mu_\Delta}{\sigma_\Delta}\right).
\end{equation}
}

\textcolor{black}{
Hence, the expectation of $Y$ is:
}

\textcolor{black}{
\begin{equation}
\label{eq:expection_Y}
\mathbb E[Y]=\mathbb P(Y=1)-\mathbb P(Y=-1)
=1-2\Phi\!\left(-\frac{\mu_\Delta}{\sigma_\Delta}\right).
\end{equation}
}

\textcolor{black}{
Let $\phi_{k\Tilde{k}}$ denote the absolute value of the expectation of $Y$. Then we have:
}

\textcolor{black}{
\begin{align}
\label{phi_en}
\phi_{k\Tilde{k}}
= \left|\mathbb{E}[Y]\right|
=\left|1-2\Phi\!\left(-\frac{\mu_{k}-\mu_{\Tilde{k}}}{\sqrt{\sigma_{k}^2+\sigma_{\Tilde{k}}^2}}\right)\right|.
\end{align}
}

\textcolor{black}{
As implied by Eq.\ref{phi_en}, $\phi_{k\Tilde{k}}$ close to $0$  typically suggests that the gray value distributions of the $k$-th class and the $\Tilde{k}$-th class have similar means, or that their within-class variances are large, leading to a statistically less stable relative thermal radiation relation between the two classes. In contrast, $\phi_{k\Tilde{k}}$ close to $1$ indicates a more pronounced separation between their distribution means with relatively small within-class variances, reflecting a more consistent and learnable thermal radiation relation pattern. Therefore, $\phi_{k\Tilde{k}}$ not only serves as a quantitative measure of the discrepancy between the two classwise gray value distributions, but also  characterizes the stability and learnability of their relative thermal radiation relation.
}

\textcolor{black}{
In practice, the parameters $\mu_k,\sigma_k$ and $\mu_{\Tilde{k}},\sigma_{\Tilde{k}}$ are often unavailable in closed form. Instead, we can approximate $\phi_{k\Tilde{k}}$ by estimating $\mathbb{E}[Y]$ from samples. Specifically, let $N_{k\Tilde{k}}$ be the number of images that contain both the $k$-th class  and the $\Tilde{k}$-th class. When $N_{k\Tilde{k}}$ is sufficiently large, the following approximation holds:
}

\textcolor{black}{
\begin{align}
\label{eq:EY_approx}
\mathbb{E}[Y] &\approx \frac{1}{N_{k\Tilde{k}}} \sum_{i=1}^{N_{k\Tilde{k}}}Y_{i},
\end{align}
}

\textcolor{black}{
\begin{align}
\label{eq:Yi_get}
Y_{i}&=\operatorname{sgn}(\mathcal{G}_{k}^{i}-\mathcal{G}_{\Tilde{k}}^{i})=\operatorname{sgn}(\mathcal{R}_{k}(\mathcal{D}_{x_{i}})-\mathcal{R}_{\Tilde{k}}(\mathcal{D}_{x_{i}})),
\end{align}
where $\mathcal{G}_{k}^{(i)}$ and $\mathcal{G}_{\Tilde{k}}^{(i)}$ denote the mean gray values of the $k$-th class  and the $\Tilde{k}$-th  class in the $i$-th infrared image, respectively. $\mathcal{R}_{k}(\mathcal{D}_{x_i})$ and $\mathcal{R}_{\Tilde{k}}(\mathcal{D}_{x_i})$ denote the position number of the $k$-th class  and the $\Tilde{k}$-th class after ranking. Finally, we obtain the statistic $\varphi_{k\Tilde{k}}$ in Section 3.2 \emph{Stability of thermal radiation relation} to approximate $\phi_{k\Tilde{k}}$ as follows:
}

\textcolor{black}{
\begin{align}
\label{eq:phi_varphi_compact}
\phi_{k\Tilde{k}}
&\approx \varphi_{k\Tilde{k}}
\triangleq
\left|\frac{1}{N_{k\Tilde{k}}}\sum_{i=1}^{N_{k\Tilde{k}}}
\operatorname{sgn}\!\Big(\mathcal{R}_{k}(\mathcal{D}_{x_i})-\mathcal{R}_{\Tilde{k}}(\mathcal{D}_{x_i})\Big)\right|. 
\end{align}
}



\section{Explanation of Equation 11}
\textcolor{black}{
Equation 11 implements a \emph{relation-aware hard-sample reweighting} principle: order relations that are currently predicted poorly should receive stronger optimization, while those already well predicted can be down-weighted to better allocate model capacity and avoid overfitting.
Concretely, we set:
\begin{equation}
\label{eq:weight_rho_rebuttal}
w_i^{\rho} = -\log\!\big(\beta\,\cdot \rho_i+1\big)+1,\qquad \beta\in(0,1),
\end{equation}
where $\rho_i=\rho(D'_x,D_x)$ measures the agreement between the predicted and reference rank orders (Spearman correlation).
This mapping enjoys the following theoretical properties.
}

\textcolor{black}{
\emph{(i) Monotone emphasis on poorly-predicted relations.}
Since
\begin{equation}
\label{eq:weight_rho_grad}
\frac{\partial w_i^{\rho}}{\partial \rho_i}
\;=\;
-\frac{\beta}{1+\beta\rho_i}
\;<\;0,
\end{equation}
$w_i^{\rho}$ decreases monotonically with $\rho_i$. Therefore, smaller $\rho_i$ (worse order prediction) yields a larger weight, which directly increases the effective gradient magnitude $w_i^{\rho}\nabla_{\theta}\mathcal{L}_{\mathrm{det}}(x_i)$ on those ``hard'' relations or samples.
}

\textcolor{black}{
\emph{(ii) Nonlinear (convex) strengthen and attenuate behavior.}
Moreover,
\begin{equation}
\label{eq:weight_rho_hess}
\frac{\partial^2 w_i^{\rho}}{\partial \rho_i^2}
\;=\;
\frac{\beta^2}{(1+\beta\rho_i)^2}
\;>\;0,
\end{equation}
so $w_i^{\rho}$ is convex in $\rho_i$, i.e., the reweighting is inherently \emph{nonlinear}.
Importantly, the sensitivity
\begin{equation}
\label{eq:weight_rho_sensitivity}
\left|\frac{\partial w_i^{\rho}}{\partial \rho_i}\right|
=
\frac{\beta}{1+\beta\rho_i}
\end{equation}
is larger when $\rho_i$ is small and decreases as $\rho_i$ grows, meaning we \emph{aggressively boost} poorly-predicted relations (low-$\rho$) while \emph{mildly suppressing} well-predicted ones (high-$\rho$), which matches our intended design rationale.
}

\textcolor{black}{
\emph{(iii) Stability-friendly bounded scaling.}
Since $\rho_i\in[-1,1]$ and $\beta\in(0,1)$, Eq.~\eqref{eq:weight_rho_rebuttal} yields a smooth and bounded scaling that avoids weight explosion and improves training stability, while still providing adaptive emphasis according to rank-order reliability.
}

\textcolor{black}{
Overall, Equation 11 can be viewed as a principled, smooth, and stable mechanism that allocates optimization budget based on the current mismatch between predicted and physical order relations, rather than a purely heuristic choice. 
}

\end{appendices}
\bibliography{sn-bibliography}

\end{document}